\documentclass{article}

\usepackage{arxiv}

\usepackage[utf8]{inputenc} 
\usepackage[T1]{fontenc}    
\usepackage{hyperref}       
\usepackage{url}            
\usepackage{booktabs}       
\usepackage{amsfonts}       
\usepackage{nicefrac}       
\usepackage{microtype}      
\usepackage{lipsum}
\usepackage{graphicx}
\usepackage{amsmath}
\usepackage{dsfont}
\usepackage{amsfonts}
\usepackage{amsmath}
\usepackage{amssymb}
\usepackage{arydshln}
\usepackage{textcomp}
\usepackage{booktabs}
\usepackage{diagbox}
\usepackage{array}
\usepackage[table,xcdraw]{xcolor}
\usepackage{booktabs}
\usepackage{scalerel}
\graphicspath{ {./images/} }

\title{Reliable Smoke Detection via Optical Flow-Guided Feature Fusion and Transformer-Based Uncertainty Modeling}

\author{\parbox{\linewidth}{\centering Nitish Kumar Mahala$^{1}$\thanks{Preprint Notice - This work has been submitted to \textit{Fire Safety Journal} for possible publication. This is the author's original manuscript and has not undergone peer review. Subsequent versions of this manuscript may differ from this one. The final published version, if accepted, will be available via Fire Safety Journal. Corresponding author: \texttt{223130002@stu.manit.ac.in}} \hfill
   Muzammil Khan $^{2}$ \hfill
   Pushpendra Kumar$^{1}$ \\[0.8em] 
  }
  \AND
  \normalfont
  $^{1}$Department of Mathematics, Bioinformatics and Computer Applications, \\ Maulana Azad National Institute of Technology Bhopal, India \\[0.5em]
  $^{2}$The Robotics and Mechatronics Group, \\University of Twente, The Netherlands  \\[0.5em]
  \texttt{223130002@stu.manit.ac.in}, \texttt{m.khan@utwente.nl}, \texttt{pkumarfma@manit.ac.in}
}

\begin{document}
\maketitle
\begin{abstract}
Fire outbreaks pose critical threats to human life and infrastructure, necessitating high-fidelity early-warning systems that detect combustion precursors such as smoke. However, smoke plumes exhibit complex spatiotemporal dynamics influenced by illumination variability, flow kinematics, and environmental noise, undermining the reliability of traditional detectors. To address these challenges without the logistical complexity of multi-sensor arrays, we propose an information-fusion framework by integrating smoke feature representations extracted from monocular imagery. Specifically, a Two-Phase Uncertainty-Aware Shifted Windows Transformer for robust and reliable smoke detection, leveraging a novel smoke segmentation dataset, constructed via optical flow-based motion encoding, is proposed. The optical flow estimation is performed with a four-color-theorem-inspired dual-phase level-set fractional-order variational model, which preserves motion discontinuities. The resulting color-encoded optical flow maps are fused with appearance cues via a Gaussian Mixture Model to generate binary segmentation masks of the smoke regions. These fused representations are fed into the novel Shifted-Windows Transformer, which is augmented with a multi-scale uncertainty estimation head and trained under a two-phase learning regimen. First learning phase optimizes smoke detection accuracy, while during the second phase, the model learns to estimate plausibility confidence in its predictions by jointly modeling aleatoric and epistemic uncertainties. Extensive experiments using multiple evaluation metrics and comparative analysis with state-of-the-art approaches demonstrate superior generalization and robustness, offering a reliable solution for early fire detection in surveillance, industrial safety, and autonomous monitoring applications.
\end{abstract}



 \section{Introduction}
Smoke detection is a pivotal task in computer vision and image processing, underpinning critical applications in fire safety, industrial automation, and surveillance. As the earliest visible manifestation of combustion, smoke enables proactive hazard mitigation, thereby reducing risks to human life and infrastructure. Conventional approaches~\cite{Cui2020} employ physical sensors such as photoelectric, ionization, carbon monoxide, and thermal detectors, which monitor particle concentrations or temperature variations. While these methods perform reliably in enclosed environments, they suffer from limited spatial coverage and an inability to capture the spatiotemporal evolution of smoke plumes in open or large-scale settings.
\par Given these constraints, vision-based techniques have emerged to exploit the rich spatial and temporal information in video streams. Yet, smoke detection remains a non-trivial challenge due to the non-deterministic dynamics of plumes, which deform under varying illumination, turbulent airflow, and heterogeneous combustion sources. Moreover, smoke shares spectral and textural properties with fog, clouds, and dust, reducing the discriminative power of conventional detection models. Recent studies~\cite{Liu2023,Jin2025} demonstrate that fusing complementary motion and appearance cues from disparate sensor modalities can mitigate these limitations. Building on this insight, the prinicpal contribution of the proposed study is a feature-level fusion framework, which integrates the following components:
\subsection{Key contributions}
\begin{enumerate}
    \item \textbf{Four-Color-theorem-inspired Dual-phase Level-set Fractional Order Variational (FCDLe-FOV) model}, incorporating a \(L_1\)-norm~\cite{Bardeji2017} data fidelity term and Marchaud fractional derivative~\cite{Muzammil2023} regularization, to accurately capture smoke motion from monocular RGB image sequences, while preserving discontinuities under non-stationary illumination.
    \item \textbf{Efficient discretization scheme for FCDLe-FOV model} by integrating Legendre-Fenchel transform-based~\cite{Chambolle2004} primal-dual algorithm with Gr\"{u}nwald-Letnikov (GL) fractional derivative~\cite{Muzammil2023}, which ensures stable convergence. 
    \item \textbf{Gaussian Mixture Model (GMM)-based~\cite{Farnoush2008} probabilistic fusion of color-encoded motion maps with RGB image features} to generate precise smoke motion segmentation masks and create a publicly available dataset \footnote{https://www.kaggle.com/datasets/nitishkumarmahala/motion-features-and-apperance-cues-datasets}. 
    \item \textbf{Two-Phase Uncertainty-Aware Shifted-windows Transformer (TP-UAST) model} that leverages the RGB images and their corresponding motion-encoded smoke masks. It is trained in two phases: Phase I optimizes detection accuracy, while Phase II jointly models aleatoric and epistemic uncertainties~\cite{Mody2024} to generate calibrated confidence estimates. 
    \item \textbf{Comprehensive experimental validation} using various evaluation metrics including accuracy, precision, recall, and F1-score~\cite{Khan2023}. Uncertainty analysis via Expected Calibration Error (ECE)~\cite{Mody2024} and reliability diagrams, present confusion matrices, uncertainty histograms, uncertainty vs. error, uncertainty by class, and plausibility analysis and confidence, as well as comparisons against state-of-the-art (SOTA) smoke detection methods.
\end{enumerate}

\subsection{\textbf{Related Work}}
Recent advances in computer vision and deep learning have significantly propelled fire safety research. However, developing robust early-warning systems remains challenging due to dynamic environments. Existing smoke detection methods, such as image-based, video-based, and optical flow information fusion-based, lack integrated feature fusion and uncertainty modeling, limiting their effectiveness in real-world hazard prevention applications.
\subsubsection{Image-based techniques}
Image-based smoke detection methods predominantly focus on extracting static spatial features such as texture, shape, color, motion patterns, and flicker signatures~\cite{Huo2022}. Traditional approaches utilize handcrafted feature extraction pipelines to distinguish smoke regions from background clutter~\cite{Vicente2002}. With the advent of deep learning, convolutional architectures such as the multi-scale dual separable convolutional neural network (CNN) proposed by Huo et al.~\cite{Huo2022}, which incorporates a CSPDarknet53 backbone with spatial pyramid pooling, have enhanced multi-scale feature representation capabilities. Similarly, Ke et al.~\cite{Gu2019} leveraged convolutional layers coupled with batch normalization to improve classification robustness, while Li et al.~\cite{Li2020} explored object detection-based CNN models, including Faster R-CNN, R-FCN, SSD, and YOLOv3, for smoke identification tasks.
\par Although these models achieve considerable performance gains, they are fundamentally constrained by their reliance on single-frame analysis, treating smoke as a purely spatial phenomenon. This absence of temporal modeling or motion analysis leads to high susceptibility to background variations and dynamic scene noise, thereby limiting their applicability in complex real-world scenarios where multi-source information, such as motion and appearance, must be jointly considered.
\subsubsection{Video-based techniques}
Video-based smoke detection methods exploit temporal consistency to enhance robustness. Lin et al.~\cite{Lin2019} combined 3D-CNN and R-CNN architectures to jointly encode both motion and appearance cues. Transformer-based frameworks have since advanced the field, including CNN-ViT hybrids~\cite{Cheng2023}, dual-branch structures~\cite{Song2024}, and Swin Transformer variant~\cite{Safarov2024} that improve multi-scale feature extraction and global context modeling. Mardani et al.~\cite{Mardani2023} demonstrated fire localization using transformer-based segmentation.
\par Despite these advancements, video-based methods typically require computationally intensive architectures and large-scale annotated datasets. Moreover, they primarily focus on feature extraction without explicitly integrating uncertainty modeling or lightweight feature fusion strategies, limiting their scalability for real-time hazard detection in dynamic environments.
\subsubsection{Optical flow information fusion-based techniques}
The dynamic motion patterns of smoke can be effectively captured using optical flow estimation~\cite{Mueller2013,Pundir2017}. Variational formulations~\cite{Kumar2016a,Huang2020} are commonly employed to solve for the optical flow field representing pixel displacements across frames. Early works applied optical flow for smoke segmentation~\cite{Mueller2013} and forest fire prediction~\cite{Pundir2017}, demonstrating its effectiveness for fluid-like objects. Khondaker et al.~\cite{khondaker2020} utilized a fire chromatic model and optical flow for robust fire segmentation, employing Mivia and Zenodo datasets for classification tasks. However, existing methods typically rely on integer-order derivatives, which assume motion continuity and struggle with capturing discontinuities inherent in turbulent or flickering smoke regions. To address these limitations, recent works~\cite{Muzammil2023,Khan2024} introduced fractional-order derivative (FOD) formulations that offered greater flexibility in capturing complex smoke motion dynamics. 
\par Building upon advancements in FOD-based optical flow approaches~\cite{Khan2024}, Khan et al.~\cite{Khan2023} proposed a CNN-based fusion framework that combines dynamical features extracted via FOD-based variational models with static appearance cues such as color, shape, and texture. Further advancements embedded active contour-based level-set segmentation~\cite{Muzammil2023} into the FOD framework, enabling precise boundary evolution critical for accurate smoke region delineation. Chunyu et al.~\cite{Chunyu2010} proposed an early video smoke detection method based on the fusion of color and motion information. Wu et al.~\cite{Wu2021} introduced an information fusion framework that integrates dense optical flow with CNN-extracted spatial features to enhance detection robustness. More recently, Kikuta et al.~\cite{Kikuta2024} developed a method that fuses optical flow variance with HSV color characteristics for daytime smoke detection.
\par Despite these advancements, existing fusion approaches lack robustness and transparency by not quantifying the predictive uncertainty, which is an essential requirement for high-reliability fire monitoring applications. These limitations underscore the need for an uncertainty-aware, robust, and generalizable information fusion framework capable of effectively handling diverse environmental conditions while ensuring reliable decision-making.
\section{Methodology}
\begin{figure}[h!]
\centering
\includegraphics[height=12.8cm, width=16cm]{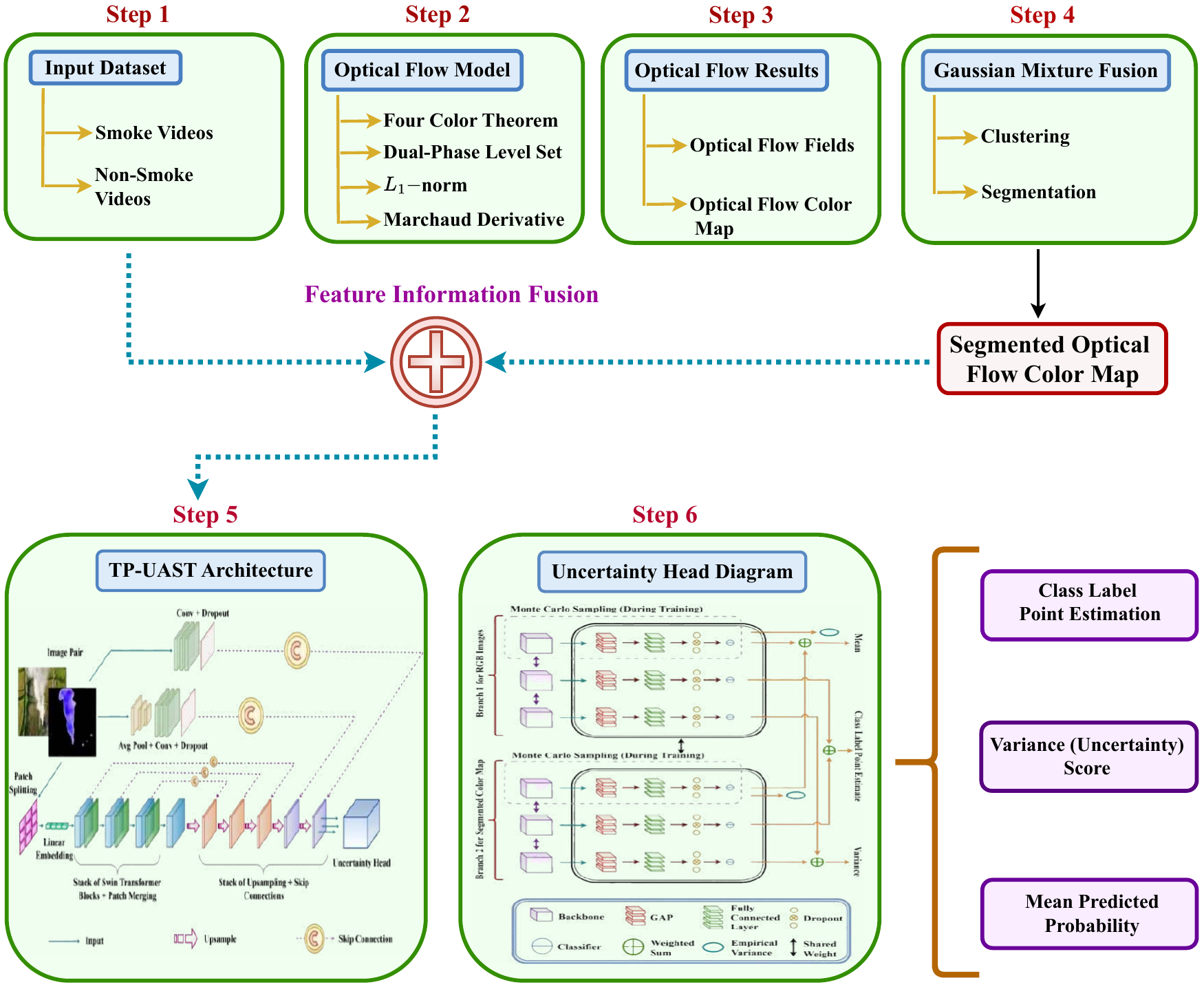}
\caption{Overview of proposed fusion framework.}\label{PMA}
\end{figure}

This study presents a robust and transparent framework for early fire prediction through accurate and reliable smoke detection by integrating fractional-order variational optical flow estimation with a novel TP-UAST model. The proposed framework comprises three primary components: fractional-order motion encoding, probabilistic motion segmentation, and uncertainty-aware classification. Optical flow estimation is formulated as a dual-phase level-set driven variational model, termed FCDLe-FOV model. The objective function incorporates an \(\mathcal{L}_{1}\)-norm-based data fidelity term~\cite{Bardeji2017} to enhance robustness against outliers and a regularization term based on the Marchaud fractional derivative~\cite{Muzammil2023}, parameterized by a smoothing coefficient \(\lambda\), to preserve motion discontinuities and complex flow boundaries. To solve the variational formulation efficiently, a Legendre-Fenchel transform-based primal-dual optimization algorithm~\cite{Chambolle2004} is employed, combined with GL discretization~\cite{Muzammil2023}, ensuring stable convergence while accurately capturing topological complexities such as triple junctions. The estimated optical flow fields are subsequently color-encoded to visualize motion dynamics and segmented using the GMM~\cite{Farnoush2008}. The GMM leverages probabilistic priors to differentiate smoke-induced motion from background dynamics, producing binary segmentation masks. These masks are then paired with the corresponding RGB frames, facilitating the construction of a high-fidelity smoke segmentation dataset that captures both motion and appearance features. For robust smoke classification, the TP-UAST model is introduced which leverages RGB images and their corresponding segmented color maps. TP-UAST has an architecture incorporating a hierarchical shifted-window self-attention mechanism to capture multi-scale spatial dependencies characteristic of smoke evolution patterns. To enhance reliability, this model is augmented with a multi-scale uncertainty estimation head that jointly models aleatoric uncertainty~\cite{Mody2024}, arising from observational noise, and epistemic uncertainty~\cite{Mody2024}, associated with model capacity limitations. The network is optimized through a two-phase learning regimen: Phase I targets smoke detection accuracy, while Phase II focuses on predictive uncertainty modeling to improve decision reliability. The complete methodology of the proposed fusion framework is illustrated in Fig.~\ref{PMA}.
\subsection{Formulation of FCDLe-FOV optical flow model}
Optical flow estimation is a fundamental task in computer vision. The objective of optical flow estimation is to examine motion patterns across successive image frames, thereby providing dynamic information about the objects in the scene. Most of the optical flow estimation methods~\cite{Muzammil2023,Khan2024} rely on minimizing the continuous energy functionals. Therefore, the proposed FCDLe-FOV model is given as,
\begin{equation}\label{6}
\begin{split}
  \mathcal{N}(\textbf{Z})= & \int_{\Omega}\bigg[\lambda\vert \mathds{I}_{t}+\nabla \mathds{I}^T\textbf{Z}\vert+\vert\vert D^{\alpha}\textbf{Z}^T\vert\vert_{F}+\nu\sum_{a=1}^{2}\left\{\vert\vert \delta_{\textbf{Z},a}^{d}\nabla\varkappa_{\textbf{Z},a} \vert\vert_{C1}\right\}\bigg]d\mathbf{X}
\end{split}
\end{equation}
where, $\nabla \mathds{I}=(\mathds{I}_{x}, \mathds{I}_{y})^T$, $\delta_{\textbf{Z},a}=\delta(\varkappa_{\textbf{Z},a})=(\delta(\varkappa_{\mathfrak{u},a}), \delta(\varkappa_{\mathfrak{v},a}))^{T}$, $\varkappa_{\textbf{Z},a}=(\varkappa_{\mathfrak{u},a},\varkappa_{\mathfrak{v},a})^T$, $||.||_{C1}$ is the sum of $L_{1}$-norm of columns in a matrix, $\vert\vert.\vert\vert_{F}$ is the Frobenius norm,  \textbf{X}=$(x,y)$, and $\textbf{Z}=(\mathfrak{u},\mathfrak{v})^T$ represents the optical flow components. Also, superscript $d$ denotes the diagonal form in a matrix. Here, $D^{\alpha}\textbf{Z}=(D_{x}^{\alpha}\textbf{Z}, D_{y}^{\alpha}\textbf{Z})^{T}$ represents the Marchaud fractional derivative of order $\alpha\in (0,1)$ and $\lambda$, $\delta$, and $\varkappa_{\textbf{Z},a}$ are the smoothing parameter, Dirac's delta function, and level curves associated with the optical flow fields $\textbf{Z}$. The positive parameter $\nu$ attracts the contour to the boundary.  
\par Now, in order to minimize the variational functional in (\ref{6}), it is decomposed based on the algorithm of Chambolle~\cite{Chambolle2004}. Hence, the proposed variational functional in (\ref{6}) is split into two parts and introduces the auxiliary variables $\hat{\textbf{Z}}=(\hat{\mathfrak{u}},\hat{\mathfrak{v}})$ that approximate $\textbf{Z}$ for a sufficiently small parameter $\theta$, which is given as
\begin{equation}\label{9}
\mathcal{N}_{\theta}(\hat{\textbf{Z}})=\int_{\Omega}\bigg[\lambda\vert \mathds{I}_{t}+\nabla \mathds{I}^T\textbf{Z}\vert+\frac{1}{2\theta}\vert\vert\hat{\textbf{Z}}-\textbf{Z}\vert\vert_{2}^2\bigg]d\mathbf{X}
\end{equation}
\begin{equation}\label{01}
\begin{split}
\mathcal{N}_{\theta}(\textbf{Z})=& \int_{\Omega}\left[\frac{1}{2\theta}\vert\vert\hat{\textbf{Z}}-\textbf{Z}\vert\vert_{2}^2+\vert\vert D^{\alpha}\textbf{Z}^T\vert\vert_{F}+\nu\left\{\sum_{a=1}^{2}\vert\vert \delta_{\textbf{Z},a}^{d}\nabla\varkappa_{\textbf{Z},a} \vert\vert_{C1}\right\}\right]d\mathbf{X}
\end{split}
\end{equation}
These convex minimization problems can be solved by using the alternating approaches as suggested by Bardeji et al.~\cite{Bardeji2017}, where at each iteration, either $\hat{\textbf{Z}}$ or $\textbf{Z}$ is updated. Specifically, the procedure is as follows: first, treat $\textbf{Z}$ as constant and solve for $\hat{\textbf{Z}}$ by minimizing $\mathcal{N}_{\theta}(\hat{\textbf{Z}})$, then, treat $\hat{\textbf{Z}}$ as constant and solve for $\textbf{Z}$ by minimizing $\mathcal{N}_{\theta}(\textbf{Z})$. Thus, in order to solve the expression in (\ref{9}), we use the primal-dual algorithm by employing the Legendre-Fenchel transform technique~\cite{Chambolle2004} and the concept of calculus of variation~\cite{Muzammil2023}.
\subsubsection{Primal-dual formulation by Legendre-Fenchel transform}
According to the Legendre-Fenchel transform definition~\cite{Chambolle2004}, we have
\begin{equation}\label{PF}
  \mbox{if } \mbox{\ \ } \phi(p)=\vert p\vert,\mbox{\ \ }\phi^{*}(d)=
\begin{cases} 
0 &, \vert d\vert\leq 1 \nonumber\\
\infty &, \vert d\vert>1
\end{cases}
, \mbox{\ }\mbox{then, we get }
\end{equation} 
\begin{equation}
\vert p\vert=\phi(p)=\sup_{\vert d\vert\leq 1}\bigg\{p\cdot d-\phi^{*}(p)\bigg\}=\sup_{\vert d\vert\leq 1}pd\nonumber
\end{equation}
where, $p=\mathds{I}_{t}+\nabla \mathds{I}^T\textbf{Z}$ . Thus, the formulation of primal-dual algorithm is given as
\begin{equation}\label{12}
  \mathcal{N}_{\theta}(\hat{\textbf{Z}})=\sup_{\vert d\vert\leq 1}\int_{\Omega}\bigg[\lambda(\mathds{I}_{t}+\nabla \mathds{I}^T\textbf{Z})d+\frac{1}{2\theta}\vert\vert\hat{\textbf{Z}}-\textbf{Z}\vert\vert_{2}^2\bigg]d\textbf{X}
\end{equation}
Now, in order to minimize the expression (\ref{12}) using the Euler-Lagrange equation, we get the following system of equations as
\begin{equation}\label{17}
  \begin{split}
   \lambda d\nabla I +\frac{1}{\theta}(\hat{\textbf{Z}}-\textbf{Z})=0
\end{split}
\end{equation}
Now, using the equation (\ref{17}) into equation (\ref{12}), we obtain
\begin{equation}\label{20}
  T(d)=\int_{\Omega}\Big[\lambda(\mathds{I}_{t}+\nabla \mathds{I}^T\textbf{Z})d-\frac{\theta}{2}\lambda^2d^2(\nabla \mathds{I}^T\nabla \mathds{I})\Big]d\mathbf{X}
\end{equation}
In order to compute the Frechet derivative of the expression in (\ref{20}), consider the functional T(d) at $d=\bar{d}+\epsilon \psi$, then, the Frechet derivative of T(d) is given by     
\begin{equation}\label{23}
 \frac{d}{d\tau}T(\bar{d})=\lambda(\mathds{I}_{t}+\nabla \mathds{I}^T\textbf{Z})-\lambda^2\theta(\nabla \mathds{I}^T\nabla \mathds{I})\bar{d}
\end{equation}
Now, using the projected gradient ascent technique in equation (\ref{23}), we get
\begin{equation}\label{24}
  \bar{d}_{temp}^{k+1}  =\bar{d}^{k}+\bigg[\lambda(\mathds{I}_{t}+\nabla \mathds{I}^T\textbf{Z})-\lambda^2\theta(\nabla \mathds{I}^T\nabla \mathds{I})\bar{d}^{k}\bigg]
\end{equation}
\begin{equation}\label{25}
\bar{d}^{k+1}=
\begin{cases}
\bar{d}_{temp}^{k+1}, & \mbox{if } \vert\bar{d}_{temp}^{k+1}\vert\leq1\\
\pm1, & \mbox{otherwise}
\end{cases}
\end{equation}
The equations (\ref{17}), (\ref{24}), and (\ref{25}) are solved alternatively to obtain $\hat{\textbf{Z}}$. The minimization of expression in (\ref{01}) is solved by again decomposing the expression into two parts, and $\hat{\textbf{Z}}$ is taken as a constant.
\begin{equation}\label{02}
\begin{split}
\mathcal{N}_{\theta}(\textbf{Z},\varkappa_{\textbf{Z},a}) = &\int_{\Omega} \bigg[\frac{1}{2\theta}\vert\vert\hat{\textbf{Z}}-\textbf{Z}\vert\vert_{2}^2+\vert\vert D^{\alpha}\textbf{Z}^T\vert\vert_{F}+\nu\sum_{a=1}^{2}\left(\delta_{\textbf{Z},a}\vert \vert\nabla \varkappa_{\textbf{Z},a}\vert\vert_{C1}\right) \bigg]d\mathbf{X}
\end{split}
\end{equation}
The connection between the four regions and the function $\textbf{Z}$ can be established by proposing the four functions such as $\textbf{Z}^{++}$, $\textbf{Z}^{+-}$, $\textbf{Z}^{-+}$, and $\textbf{Z}^{--}$ according to the Four-Color theorem~\cite{vese2002}. These functions effectively restrict $\textbf{Z}$ to each of the four regions, as outlined below
$$\textbf{Z}(x,y)=\textbf{Z}^{ij}$$
\begin{equation}\label{PW}
 \mbox{where },  ij=
  \begin{cases}\nonumber
    ++, & \mbox{if } (r,s,n\Delta\tau)\vert\varkappa_{\mathfrak{u},1,r,s}^{(n)}>0 \mbox{ on } \varkappa_{\mathfrak{u},2,r,s}^{(n)}>0 \\
    +-, & \mbox{if } (r,s,n\Delta\tau)\vert\varkappa_{\mathfrak{u},1,r,s}^{(n)}>0 \mbox{ on } \varkappa_{\mathfrak{u},2,r,s}^{(n)}<0\\
    -+, & \mbox{if } (r,s,n\Delta\tau)\vert\varkappa_{\mathfrak{u},1,r,s}^{(n)}<0 \mbox{ on } \varkappa_{\mathfrak{u},2,r,s}^{(n)}>0 \\
    --, & \mbox{if } (r,s,n\Delta\tau)\vert\varkappa_{\mathfrak{u},1,r,s}^{(n)}<0 \mbox{ on } \varkappa_{\mathfrak{u},2,r,s}^{(n)}<0
  \end{cases}
\end{equation}
The relation between $\textbf{Z}$, four functions $\textbf{Z}^{++}$, $\textbf{Z}^{+-}$, $\textbf{Z}^{-+}$, and $\textbf{Z}^{--}$, and the level set functions $\varkappa_{1}$ and $\varkappa_{2}$ can again be expressed using the function $\mathcal{H}$ as
\begin{equation}
  \textbf{Z}=\textbf{Z}^{++}\mathcal{H}_{\textbf{Z}_{1}}+\textbf{Z}^{+-}\mathcal{H}_{\textbf{Z}_{2}}+\textbf{Z}^{-+}\mathcal{H}_{\textbf{Z}_{3}}+\textbf{Z}^{--}\mathcal{H}_{\textbf{Z}_{4}}\nonumber
\end{equation}
where, $\mathcal{H}_{\textbf{Z}_{1}}=(\mathcal{H}_{\mathfrak{u},1}\mathcal{H}_{\mathfrak{u},2},\mathcal{H}_{\mathfrak{v},1}\mathcal{H}_{\mathfrak{v},2})^T, \mathcal{H}_{\textbf{Z}_{2}}=(\mathcal{H}_{\mathfrak{u},1}(I-\mathcal{H}_{\mathfrak{u},2}), \mathcal{H}_{\mathfrak{v},1}(I-\mathcal{H}_{\mathfrak{v},2}))^T, \mathcal{H}_{\textbf{Z}_{3}}=((I-\mathcal{H}_{\mathfrak{u},1})\mathcal{H}_{\mathfrak{u},2},(I-\mathcal{H}_{\mathfrak{v},1})\mathcal{H}_{\mathfrak{v},2})^T,$
$\mathcal{H}_{\textbf{Z}_{4}}=((I-\mathcal{H}_{\mathfrak{u},1})(I-\mathcal{H}_{\mathfrak{u},2}),(I-\mathcal{H}_{\mathfrak{v},1})(I-\mathcal{H}_{\mathfrak{v},2}))^T$. Thus, in accordance with the Vese-Chan model~\cite{vese2002}, the expression (\ref{02}) can be written as 
\begin{align}\label{03}
 \mathcal{N}(\textbf{Z}^{ij},\varkappa_{\textbf{Z},a}) =\int_{\Omega}\sum_{i,j} \left(\frac{1}{2\theta}\vert\vert\hat{\textbf{Z}}-\textbf{Z}^{ij}\vert\vert_{2}^2+\vert\vert D^{\alpha}\textbf{Z}^{ij}\vert\vert_{F}\right)\mathcal{H}^{i}_{\varkappa_{\textbf{Z},a}}\mathcal{H}^{j}_{\varkappa_{\textbf{Z},a}}+\nu\left\{\sum_{a=1}^{2}\left(\delta_{\textbf{Z},a}\vert\vert \nabla \varkappa_{\textbf{Z},a}\vert\vert_{C1}\right)\right\}d\mathbf{X}
\end{align}
where $i, j \in \{+,-\}$, and $\mathcal{H}^{+}_{\varkappa_{\textbf{Z},a}}= \mathcal{H}_{\varkappa_{\textbf{Z},a}} $, $\mathcal{H}^{-}_{\varkappa_{\textbf{Z},a}} = 1-\mathcal{H}_{\varkappa_{\textbf{Z},a}} $ for $a = 1,2$. So, the Euler-Lagrange equations derived from minimizing the variational functional (\ref{03}) using the calculus of variations are given as
\begin{align}\label{F1}
\frac{\partial \varkappa_{\textbf{Z},a}}{\partial \tau}  = -\delta_{\textbf{Z},a} & \left[\frac{1}{2\theta} \sum_{i,j} j\cdot \left[(\hat{\textbf{Z}}-\textbf{Z}^{ij})^2 + \vert\vert D^{\alpha}\textbf{Z}^{ij}\vert\vert_{F}\right] -\nu \nabla\cdot\left(\frac{\nabla\varkappa_{\textbf{Z},a}}{\vert\nabla\varkappa_{\textbf{Z},a}\vert}\right)\right]
\end{align}
\begin{equation}\label{F3}
  \textbf{Z}^{ij}=\hat{\textbf{Z}}-2\theta\{\left(D_{-}^{\alpha}D_{+}^{\alpha}\right)^T e\}\textbf{Z}^{ij}\mbox{over } (x,y,\tau)
\end{equation}
\normalsize
where, $\partial\tau$ denotes the artificial time step in which the level set functions evolve. These equations allow to compute $\textbf{Z}^{ij}$ over the auxiliary flow field $\hat{\textbf{Z}}$.
\subsection{Numerical discretization and solution}
\subsubsection{Dual-phase level set discretization scheme}
We approximate Heaviside's unit step function and Dirac's delta function for numerical implementation purposes as 
\begin{align}
&\mathcal{H}(\varkappa)\approx \mathcal{H}_{\epsilon}(\varkappa)=\frac{1}{2}\left[1+\frac{2}{\pi}\arctan\left(\frac{\varkappa}{\epsilon}\right)\right] \mbox{\ \ \ \ and \ \ \ \ \ \ }\delta(\varkappa)\approx \delta_{\epsilon}(\varkappa)=\frac{1}{\pi}\frac{\epsilon}{\epsilon^2+\varkappa^2}\nonumber
\end{align}
Let $\varkappa_{\textbf{Z},1,r,s}^{0}$ and $\varkappa_{\textbf{Z},2,r,s}^{0}$ represent the initial approximations of the level surfaces, and $\varkappa_{\textbf{Z},1,r,s}^{(n)}$ and $\varkappa_{\textbf{Z},2,r,s}^{(n)}$ denote the $n^{th}$ iterate approximations of the level surfaces. Then, the two-phase level set discretization is performed according to the theory of Vese and Chan~\cite{vese2002} is given as
\begin{equation}\label{N1}
  \frac{\partial \varkappa_{\textbf{Z},a}(x,y,\tau)}{\partial \tau}\approx\frac{\varkappa_{\textbf{Z},a,r,s}^{(n+1)}-\varkappa_{\textbf{Z},a,r,s}^{(n)}}{\Delta\tau} 
\end{equation}
\begin{align}\label{N2}
  \nabla\cdot\left(\frac{\nabla \varkappa_{\textbf{Z},a}}{\vert\nabla \varkappa_{\textbf{Z},a}\vert}\right)=C_{1}\left(\varkappa_{\textbf{Z},a,r+1,s}^{(n)}-\varkappa_{\textbf{Z},a,r,s}^{(n+1)}\right)+&C_{2}\left(\varkappa_{\textbf{Z},a,r-1,s}^{(n)}-\varkappa_{\textbf{Z},a,r,s}^{(n+1)}\right)+C_{3}\left(\varkappa_{\textbf{Z},a,r,s+1}^{(n)}-\varkappa_{\textbf{Z},a,r,s}^{(n+1)}\right)\nonumber\\
  &+C_{4}\left(\varkappa_{\textbf{Z},a,r,s-1}^{(n)}-\varkappa_{\textbf{Z},a,r,s}^{(n+1)}\right)
\end{align}
\footnotesize{
\begin{align}\label{N3}
\mbox{where, }
&C_{1}=\left\{\left(\frac{\varkappa_{\textbf{Z},k,r+1,s}^{(n)}-\varkappa_{\textbf{Z},k,r,s}^{(n)}}{h}\right)^2+\left(\frac{\varkappa_{\textbf{Z},k,r,s+1}^{(n)}-\varkappa_{\textbf{Z},k,r,s-1}^{(n)}}{2h}\right)^2\right\}^{-\frac{1}{2}},\nonumber\\
&C_{2}=\left\{\left(\frac{\varkappa_{\textbf{Z},k,r,s}^{(n)}-\varkappa_{\textbf{Z},k,r-1,s}^{(n)}}{h}\right)^2+\left(\frac{\varkappa_{\textbf{Z},k,r-1,s+1}^{(n)}-\varkappa_{\textbf{Z},k,r-1,s-1}^{(n)}}{2h}\right)^2\right\}^{-\frac{1}{2}}  \nonumber \\
&C_{3}=\left\{\left(\frac{\varkappa_{\textbf{Z},k,r+1,s}^{(n)}-\varkappa_{\textbf{Z},k,r-1,s}^{(n)}}{h}\right)^2+\left(\frac{\varkappa_{\textbf{Z},k,r,s+1}^{(n)}-\varkappa_{\textbf{Z},k,r,s}^{(n)}}{2h}\right)^2\right\}^{-\frac{1}{2}},\nonumber\\
&C_{4}=\left\{\left(\frac{\varkappa_{\textbf{Z},k,r+1,s}^{(n)}-\varkappa_{\textbf{Z},k,r-1,s}^{(n)}}{h}\right)^2+\left(\frac{\varkappa_{\textbf{Z},k,r,s}^{(n)}-\varkappa_{\textbf{Z},k,r,s-1}^{(n)}}{2h}\right)^2\right\}^{-\frac{1}{2}}\nonumber 
\end{align}}
\normalsize
\subsubsection{Fractional derivative discretization scheme}
Let the dimensions of the optical flow components $\textbf{Z}=(\mathfrak{u,v})^T$ be same as those of the reference image of size $m\times n$. Now, by using the GL derivative~\cite{Muzammil2023}, we discretize the optical flow fields $\textbf{Z}^{ij}$ as 
\begin{equation}\label{uf4}
  \{(D_{-}^{\alpha}D_{+}^{\alpha})e\}\textbf{Z}^{ij}\approx \sum_{(\bar{r},\bar{s})\in\xi(r,s)}w_{q_{\bar{r}\bar{s}}}^{\alpha}\{\textbf{Z}^{ij}-\bar{\textbf{Z}}^{ij}\}
\end{equation}
Here, the set $\xi$ represents all the pixels in the neighborhood of the pixel location $(r,s)$ in both the $x$ and $y$ directions, where $q_{\bar{r}\bar{s}}=\max\left[\vert\bar{r}-r\vert,\vert\bar{s}-s\vert\right]$. \\
On using the equations (\ref{N1}), (\ref{N2}), and (\ref{uf4}), we get the following discretized system of equations for optical flow $\textbf{Z}$ as
\begin{equation}\label{uf6}
\textbf{Z}_{r,s}^{(n+1),ij}=\mathfrak{R}^{-1}\left[\hat{\textbf{Z}}_{r,s}^{(n)}+2\theta\sum_{(\bar{r},\bar{s})\in\xi(r,s)}w_{q_{\bar{r}\bar{s}}}^{\alpha}\textbf{Z}_{\bar{r},\bar{s}}^{(n),ij}\right]
\end{equation}
here, $\textbf{Z}^{ij}(\bar{r},\bar{s})=\textbf{Z}_{\bar{r},\bar{s}}^{ij}$, and $\mathfrak{R}=1+2\theta\sum_{(\bar{r},\bar{s})\in\xi(r,s)}w_{q_{\bar{r}\bar{s}}}$. 
\footnotesize{
\begin{align}\label{uf10}
  \varkappa_{\textbf{Z},a,r,s}^{(n+1)}= & \frac{1}{C_{a}^{(n)}}\left[\varkappa_{\textbf{Z},a,r,s}^{(n)}+\gamma_{a}^{(n)}\left\{C_{1}^{n}\varkappa_{\textbf{Z},a,r+1,s}^{(n)}+C_{2}^{n}\varkappa_{\textbf{Z},a,r-1,s}^{(n)}+C_{3}^{n}\varkappa_{\textbf{Z},a,r,s+1}^{(n)}+C_{4}^{n}\varkappa_{\textbf{Z},a,r,s-1}^{(n)}\right\}\right.\nonumber\\
  &\left.-\frac{\gamma_{a}^{(n)}}{2}\left\{\frac{1}{2\theta}\left\{ \sum_{i} i(\hat{\textbf{Z}}^{(n)}-\textbf{Z}^{(n),i+})^{d} (\hat{\textbf{Z}}^{(n)}-\textbf{Z}^{(n),i+})\right\}+(h^{-\alpha})^2\right.\right.\nonumber\\
       & \left.\left.\left\{\sum_{i} i \mbox{ }diag \left(\sum_{q=0}^{W}w_{q}^{(\alpha)}E_{\textbf{x}}^{-q}\textbf{Z}^{i+,T}\right)^T \left(\sum_{q=0}^{W}w_{q}^{(\alpha)}E_{\textbf{x}}^{-q}\textbf{Z}^{p+,T}\right)  \right\}\right\}\mathcal{H}(\varkappa_{\textbf{Z},1}^{(n)})\right. \nonumber\\
       &\left.-\frac{\gamma_{a}^{(n)}}{2}\left\{\frac{1}{2\theta}\left\{\sum_{p} p(\hat{\textbf{Z}}^{(n)}-\textbf{Z}^{(n),p-})^{d} (\hat{\textbf{Z}}^{(n)}-\textbf{Z}^{(n),p-})\right\}+(h^{-\alpha})^2\right.\right.\nonumber\\
    &\left.\left. \left\{\sum_{i} i \mbox{ }diag \left(\sum_{q=0}^{W}w_{q}^{(\alpha)}E_{\textbf{x}}^{-q}\mathcal{Z}^{i-,T}\right)^T \left(\sum_{q=0}^{W}w_{q}^{(\alpha)}E_{\textbf{x}}^{-q}\textbf{Z}^{i-,T}\right)\right\}\right\}(2-\mathcal{H}(\varkappa_{\textbf{Z},2}^{(n)})\right]\nonumber\\
\end{align}}
\normalsize
\noindent where, $\gamma_{a}^{(n)}=\frac{\Delta\tau}{h^2}\delta_{\epsilon}(\varkappa^{(n)}_{a}(r,s))$ and $C^{(n)}_{a}=1+\gamma_{a}^{(n)}(C_{1}+C_{2}+C_{3}+C_{4})$ for $a=1$ and $2$. Hence, these equations produce the optical flow fields. In this work, the color maps represent the estimated optical flow for smoke as well as non-smoke images, which are further processed to create the segmented binary mask using GMM technique.
\subsection{Stability and convergence analysis}
\subsubsection{Stability analysis}
In order to analyze the stability of the discretized system of equations (\ref{uf6}). The fourier decomposition of $\textbf{Z}^{(n),ij}_{r,s}$ can be expressed as $\textbf{Z}^{(n),ij}_{r,s}=\textbf{Z}^{(n),ij}e^{i(kr+ls)}$. Then, the expression in (\ref{uf6}) becomes 
\begin{align}\label{SA}
  \textbf{Z}^{(n+1),ij}_{r,s} &=\mathfrak{R}^{-1}\left[\textbf{Z}_{r,s}^{(n),ij}+2\theta\sum_{(\bar{r},\bar{s})\in\xi(r,s)}w_{q_{\bar{r}\bar{s}}}^{\alpha}\textbf{Z}_{r,s}^{(n),ij}e^{i(k\bar{r}+l\bar{s})}\right] \nonumber\\
   & =\mathfrak{R}^{-1}\left[1+2\theta\sum_{(\bar{r},\bar{s})\in\xi(r,s)}w_{q_{\bar{r}\bar{s}}}^{\alpha}e^{i(k\bar{r}+l\bar{s})}\right]\textbf{Z}_{r,s}^{(n),ij} \nonumber\\
   & =G(k,l)\textbf{Z}_{r,s}^{(n),ij}
\end{align} 
here, $G(k,l)$ represents the amplification factor of the expression in (\ref{uf6}), and $k$ and $l$ are the wavenumbers. Hence, the amplification factor is written as  
 \[ 
G(k,l)=\mathfrak{R}^{-1}\left[1+2\theta\sum_{(\bar{r},\bar{s})\in\xi(r,s)}w_{q_{\bar{r}\bar{s}}}^{\alpha}e^{i(k\bar{r}+l\bar{s})}\right]
  \]\\
Now, 
\begin{align}\label{SA2}
\vert G(k,l)\vert= & \Big| \mathfrak{R}^{-1}\Big[1+2\theta\sum_{(\bar{r},\bar{s})\in\xi(r,s)}w_{q_{\bar{r}\bar{s}}}^{\alpha}e^{i(k\bar{r}+l\bar{s})}\Big]\Big|\\
&\leq \Big| \mathfrak{R}^{-1}\Big|\Big|\Big[1+2\theta\sum_{(\bar{r},\bar{s})\in\xi(r,s)}w_{q_{\bar{r}\bar{s}}}^{\alpha}e^{i(k\bar{r}+l\bar{s})}\Big]\Big|\nonumber
\end{align}
Since $|e^{i(k\bar{r}+l\bar{s})}|=1$, then we get
\begin{equation}\label{SA3}
\Big|\sum_{(\bar{r},\bar{s})\in\xi(r,s)}w_{q_{\bar{r}\bar{s}}}^{\alpha}e^{i(k\bar{r}+l\bar{s})}\Big|\leq \sum_{(\bar{r},\bar{s})\in\xi(r,s)}|w_{q_{\bar{r}\bar{s}}}^{\alpha}|
\end{equation}
Therefore, according to theorem~\cite{Neelan2023}, we obtain
\begin{equation}\label{SA4}
| \mathfrak{R}^{-1}|[1+2\theta\sum_{(\bar{r},\bar{s})\in\xi(r,s)}|w_{q_{\bar{r}\bar{s}}}^{\alpha}|]\leq1
\end{equation}
On simplifying the expression in (\ref{SA4}), we have 
\begin{equation}\label{SA5}
1+2\theta\sum_{(\bar{r},\bar{s})\in\xi(r,s)}|w_{q_{\bar{r}\bar{s}}}^{\alpha}|\leq \mathfrak{R}
\end{equation}
Thus, the proposed FCDLe-FOV model is stable, when $\sum_{(\bar{r},\bar{s})\in\xi(r,s)}|w_{q_{\bar{r}\bar{s}}}^{\alpha}|\leq \frac{\mathfrak{R}-1}{2\theta}$.
\subsubsection{Convergence analysis}
Let the sequence $\{\textbf{Z}^{(n),ij}_{r,s}\}$ has an error term $e^{(n),ij}_{r,s}$ such that $\lim_{n\to\infty}e^{(n),ij}_{r,s}=0$ with $|G(k,l)|\leq1$, then,
\[
e_{r,s}^{(n+1),ij} = G(k,l) e_{r,s}^{(n),ij} \mbox{\ \ }\mbox{as given in expression (\ref{SA}) }.
\]
Now,
\[
\left| \textbf{Z}_{r,s}^{(n+1),ij} - \textbf{Z}_{r,s}^{(n),ij} \right| = \left| G(k,l) e_{r,s}^{(n),ij} - e_{r,s}^{(n),ij} \right| = e_{r,s}^{(n),ij} \left| G(k,l) - 1 \right|
\]
Since \( \lim\limits_{n \to \infty} e_{r,s}^{(n),ij} = 0 \) and \( |G(k,l) - 1| \) is bounded, therefore
\[
\lim\limits_{n \to \infty} \left| \textbf{Z}_{r,s}^{(n+1),ij} - \textbf{Z}_{r,s}^{(n),ij} \right| = 0
\]
Hence, $\{\textbf{Z}^{(n),ij}_{r,s}\}$ is convergent.

\subsection{GMM-based binary mask for information fusion}
\begin{figure}[h!]
\begin{center}
\includegraphics[width=14.5cm, height=7cm]{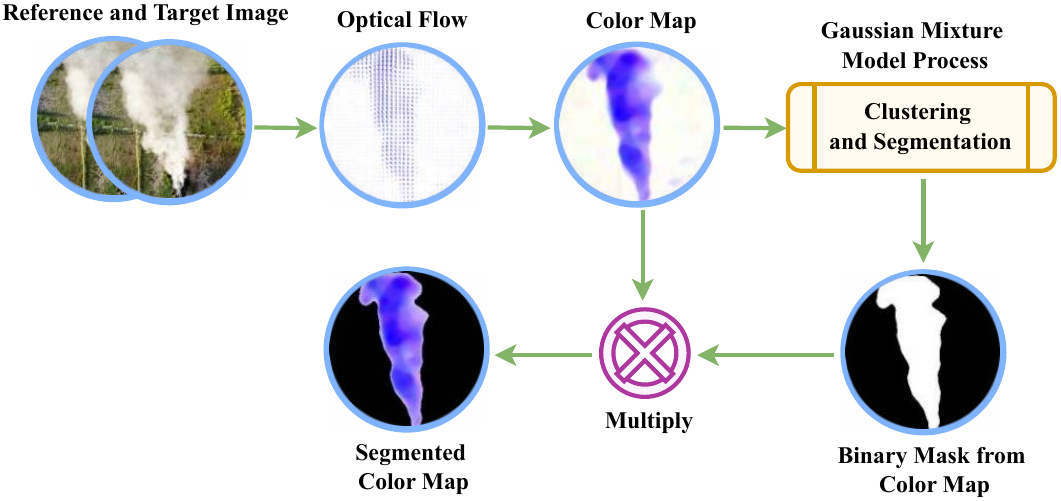}\\
\caption{Extraction of smoke motion features using optical flow color maps.}
\label{BM}
\end{center}
\end{figure}

The segmentation pipeline employed to produce the binary mask is shown in Fig.~\ref{BM}. To identify the smoke region of interest (RoI), the system utilizes a dense optical flow color map that encodes motion information from the flow field. However, the spatiotemporal complexity of smoke motion leads to uneven and dispersed pixel intensity values in the color map, influenced by diverse motion dynamics and background noise. To address this, a GMM~\cite{Farnoush2008} technique is used to segment the motion map. GMM is a probabilistic clustering method that represents data as a combination of multiple gaussian distributions, allowing it to effectively handle complex intensity patterns. This segmentation facilitates the extraction of motion features, which are then used in the subsequent step.
\section{Architecture of proposed TP-UAST model}
\begin{figure}[h!]
\centering
\begin{tabular}{cc}
\begin{minipage}[t]{0.55\textwidth}
\centering
\textbf{(a)} \\
\includegraphics[width=\textwidth]{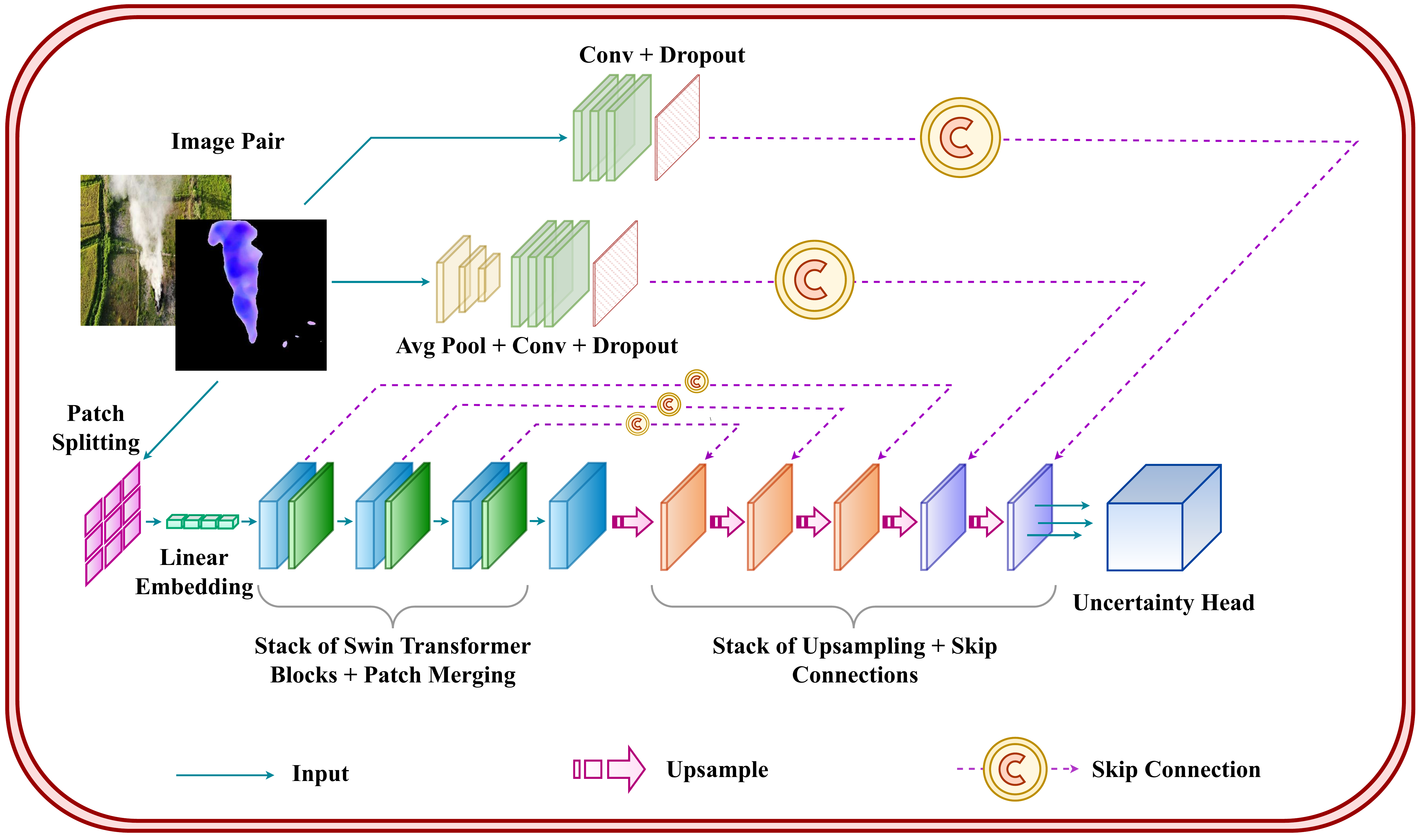}
\end{minipage}
&
\begin{minipage}[t]{0.4\textwidth}
\centering
\textbf{(b)} \\
\includegraphics[width=\textwidth]{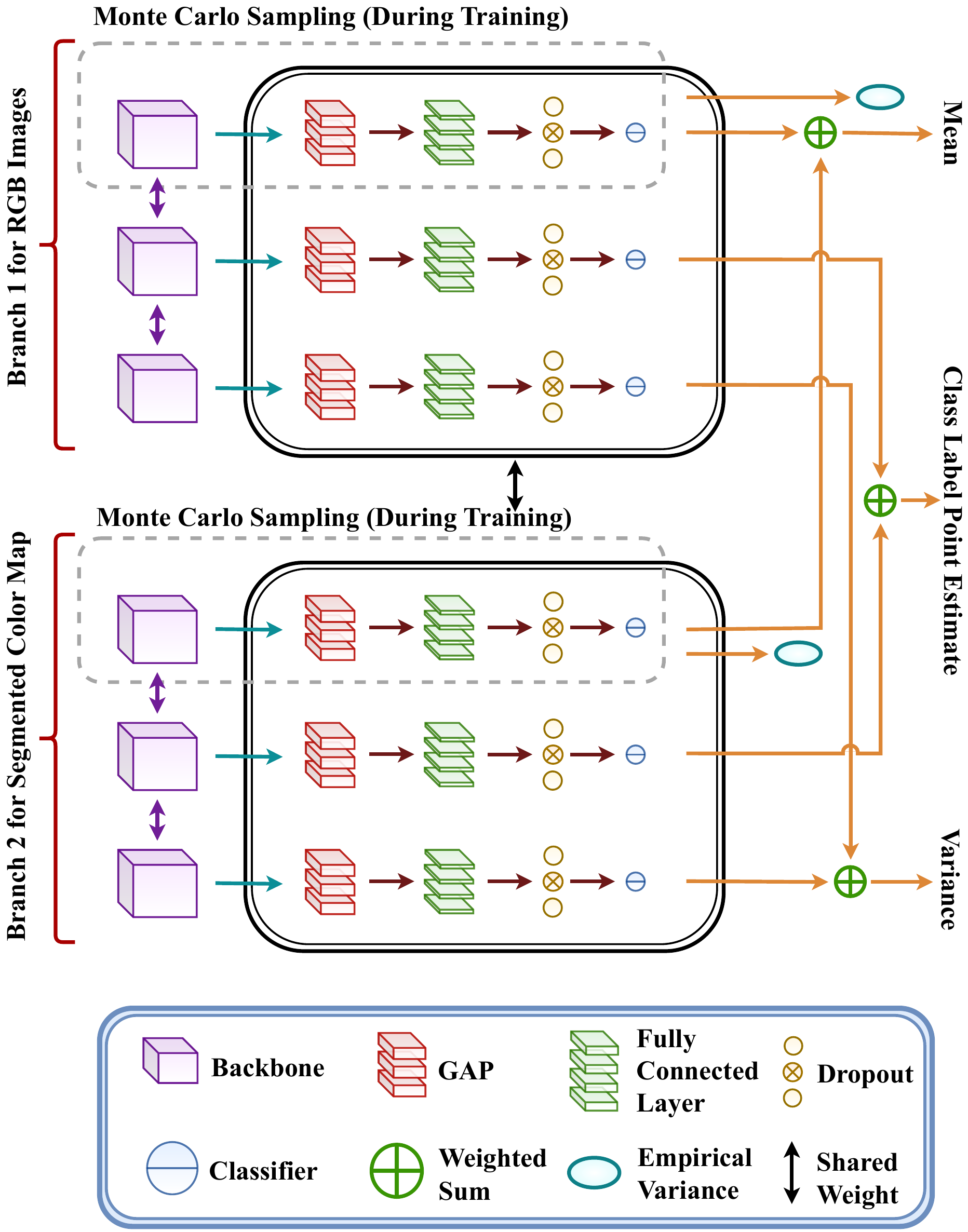}
\end{minipage}
\end{tabular}
\caption{(a) The Swin Transformer backbone; (b) The uncertainty-aware prediction head.}
\label{UAST}
\end{figure}
The architecture of the proposed TP-UAST model is illustrated in Fig.~\ref{UAST}. It integrates an uncertainty-aware prediction head into a shared Swin Transformer backbone~\cite{liu2021}, leveraging its hierarchical, window-based self-attention~\cite{Vaswani2017} to effectively capture multi-scale spatial features. TP-UAST processes two parallel input branches, one for the RGB image and one for its corresponding segmented color map, each passed independently through the same backbone and decoder modules. To enable reliable uncertainty estimation for the predicted class labels, the model jointly captures epistemic and aleatoric uncertainties. The epistemic uncertainty is optimized via Monte Carlo (MC) sampling with random input transformations, including horizontal flipping and rotations, while ensuring consistency between paired branches by using the same random seed. Aleatoric uncertainty is captured through a dedicated uncertainty-aware prediction head that regresses the predicted mean and variance along with the point prediction. This design enhances both discriminative accuracy and prediction confidence under challenging smoke detection conditions. The core architecture following the input step is as follows:
\begin{itemize}
 \item [$\bullet$]\textbf{Patch splitting and linear embedding:} The TP-UAST model pipeline begins with the patch splitting layer~\cite{dong2023}, which partitions the input sample into non-overlapping patches of size $4 \times 4$ pixels. Each patch is subsequently linearly embedded~\cite{Raisi2020} into a feature vector of dimension $96$, resulting in a sequence of embeddings. This operation effectively reduces the spatial resolution of the input by a factor of $4$, while preserving critical spatial and structural information for downstream processing. \vspace{5pt}
 \item [$\bullet$]\textbf{Swin Transformer blocks:} The network comprises four hierarchical stages to refine the feature representations at progressively larger scales, with each stage consisting of multiple Swin Transformer blocks. At the end of each stage, except the final one, patch merging layers~\cite{liu2021} are employed to downsample the feature maps by a factor of two while doubling the number of feature channels. In the initial stage, two Swin Transformer blocks operate with four attention heads and a window size of $7\times7$, processing features at a dimension of $96$. The second stage also includes two blocks with four attention heads but doubles the feature dimension to $192$ while maintaining the same window size. Stage three expands to four blocks with eight attention heads, further increasing the feature dimension to $384$, allowing the model to capture more intricate spatial patterns. Finally, the fourth stage concludes the refinement process with two blocks and eight attention heads, pushing the feature dimension to $768$. This structure progressively reduces the spatial resolution and increases the feature dimensionality, allowing the model to focus on increasingly abstract and high-level features.
 \item [$\bullet$]\textbf{Decoder blocks and skip connections:} Following the transformer blocks, TP-UAST incorporates five sequential decoder blocks to restore the original spatial dimensions of the input, while integrating feature information from earlier stages of the network. Each decoder block begins by upsampling the input feature map through bilinear interpolation, doubling the spatial dimensions. The upsampled feature map is then concatenated with the corresponding feature map from the encoder through skip connections, ensuring that both high-level semantic features and low-level spatial details are preserved. In each decoder block, after concatenation, the feature map is further processed through a series of two Conv layers, each followed by a ReLU activation. It refines the feature representations post-upsampling. The decoder blocks are applied sequentially, beginning with the first block that upsamples the feature maps from Swin Transformer block $4$ and combines them with those from block $3$. The subsequent blocks progressively upsample the feature maps and merge them with the outputs from Swin Transformer blocks $2$ and $1$, respectively. The fourth and fifth decoder blocks perform additional upsampling, combining the feature maps with the original convolved input features to ensure the preservation of fine spatial details.
 \end{itemize}
 \subsection{Uncertainty-aware prediction head}
 The uncertainty-aware prediction head operates on the final feature maps extracted from each input branch, which comprises three parallel sub-heads based on a shared layer design. Each sub-head first applies global average pooling (GAP) to reduce spatial dimensions, followed by a fully connected layer with dropout regularization and a final classification layer. The primary head outputs the predicted class score, while two auxiliary heads predict the mean and variance of the predicted probability distribution corresponding to the input sample, respectively. During training, the empirical mean and standard deviation computed via MC sampling serve as target signals for the mean and variance prediction sub-heads, respectively.
 \section{Two-Phase training procedure}
 The TP-UAST model is trained on pairs of RGB images and their corresponding segmented color maps, resized to \(256\times256\) pixels, using a two-phase curriculum. This approach is designed to decouple point and mean prediction from variance learning for the class of the input sample. In phase I, only the primary classification and mean prediction sub-heads, along with the backbone weights, are trained. The variance sub-head remains frozen in this phase. A composite loss is optimized that balances the binary cross-entropy on both the point prediction and the mean prediction. The Adam optimizer is used with an initial learning rate of \(1\times10^{-4}\), decayed by a polynomial schedule. Training continues until the epoch-averaged loss falls below a threshold value of $0.2$, at which point phase I completes. During phase II, all weights of the model except those of the variance sub-head are frozen. In this phase, the $L_{1}$ loss is computed between the output $\sigma^{2}$ of the variance sub-head and the empirical variance \(\hat\sigma^{2}_{\mathrm{MC}}\) computed via MC sampling. This targeted regression teaches the variance head to match the observed spread of predictions.
 \section{Experimental results and discussion}
 \subsection{Performance evaluation metrics}
The performance of the FCDLe-FOV model is assessed using three key metrics: Average Angular Error (AAE), which measures the angular deviation between estimated and ground truth flow vectors; Average Endpoint Error (AEPE), which calculates the Euclidean distance between the predicted and ground truth flow vectors; and Average Error Normal to the Gradient (AENG), which evaluates the accuracy of flow estimation in directions perpendicular to image gradients, particularly around object boundaries~\cite{Khan2024}. Moreover, a Structural Similarity Index Measure (SSIM), which quantifies the structural similarity between the optical flow color maps of the reference and noisy images, is employed for robustness evaluation.
\par The TP-UAST framework is evaluated across four complementary dimensions: Discriminative Performance, via accuracy, precision, recall and \(F_{1}\)-score to quantify classification correctness and positive-class sensitivity; Calibration, via reliability diagram to ensure predicted probabilities align with empirical frequencies; Predictive Uncertainty Quantification, where histograms of per-sample predictive standard deviations assess uncertainty distribution, scatter plots of uncertainty versus absolute error to validate error-uncertainty correlation, and class-stratified boxplots to detect systematic biases; and Plausibility Analysis, with empirical histograms of Z-score \(Z=(\ell-\mu)/\sigma\) and plausibility confidences \(C=\exp(-\tfrac12Z^2)\), which exhibit a $Z$ and a $C$ value corresponding to each input's posterior predictive distribution \(\ell\sim\mathcal{N}(\mu,\sigma^{2})\). Here \(\ell\), \(\mu\) and \(\sigma^{2}\) denote the class label point estimate, predicted probability, and uncertainty score, respectively.
\subsection{Experimental discussion}
The experiments conducted in this study have been performed using MATLAB R2023a and WSL Ubuntu 22.04, running on a Windows 11 system equipped with an NVIDIA GeForce RTX 4080 Laptop GPU. The experimental design consists of three main components: the FCDLe-FOV model, used for estimating robust smoke motion features; a GMM technique, for segmenting the smoke RoI; and the TP-UAST deep learning model, trained on fusion datasets. Furthermore, the plausibility of the TP-UAST model is thoroughly validated through quantitative and qualitative analyses.    
\begin{figure}[h!]
\begin{center}
\begin{tabular}{ccccc}
\textbf{Smoke1}&\textbf{Smoke2}&\textbf{Smoke3}&\textbf{Smoke4}&\textbf{Smoke5}\\
\includegraphics[width=2.8cm, height=2.8cm]{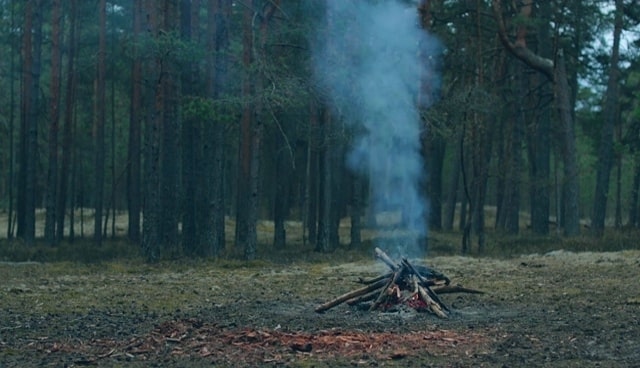} &\includegraphics[width=2.8cm, height=2.8cm]{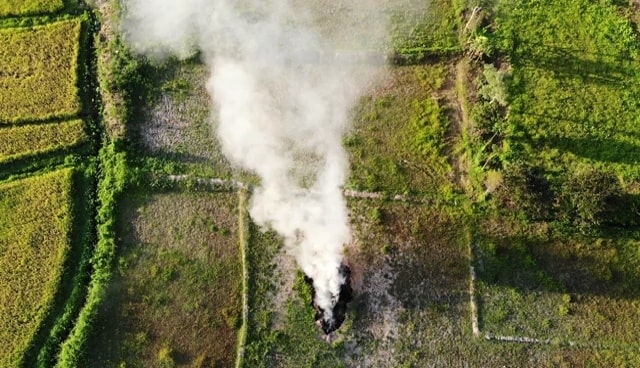}&\includegraphics[width=2.8cm, height=2.8cm]{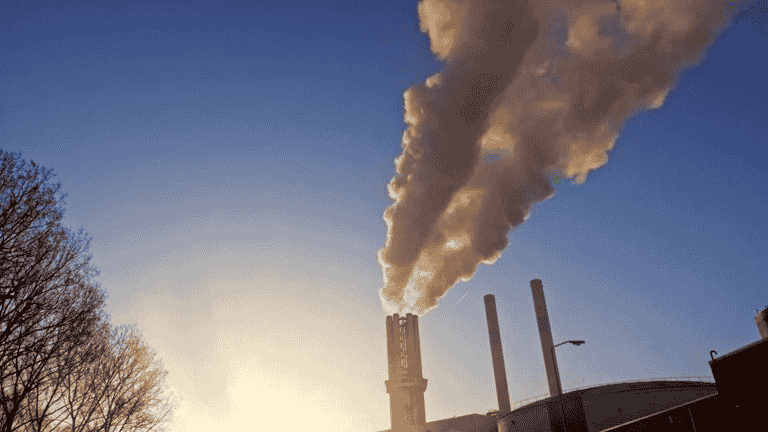}&\includegraphics[width=2.8cm, height=2.8cm]{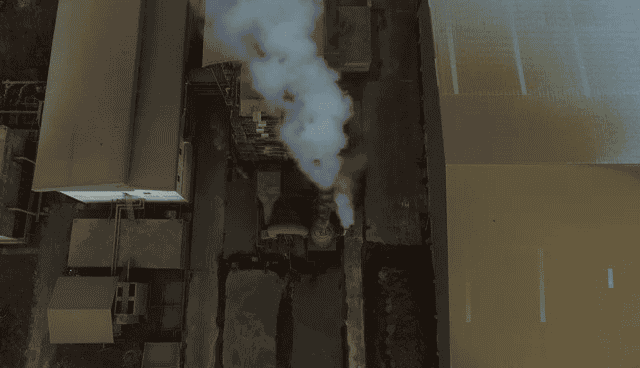}&\includegraphics[width=2.8cm, height=2.8cm]{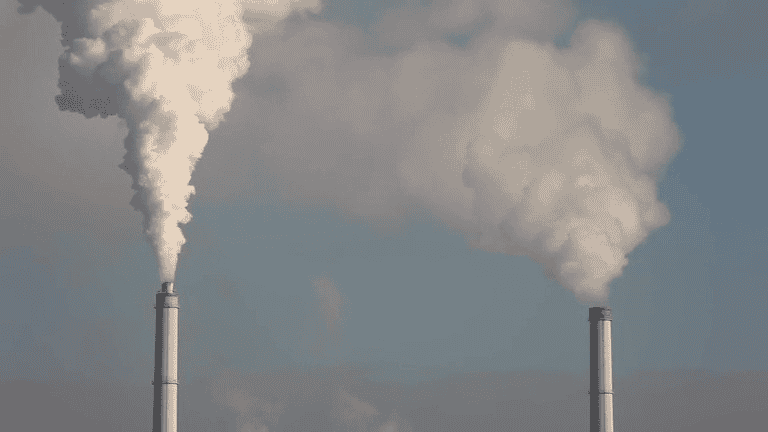}\\
\\
\fbox{\includegraphics[width=2.5cm, height=2.5cm]{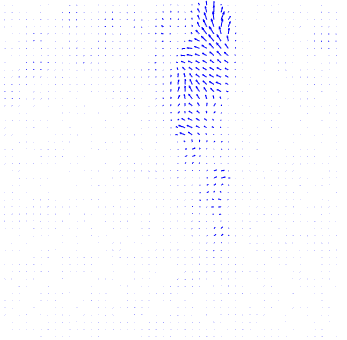}} &\fbox{\includegraphics[width=2.5cm, height=2.5cm]{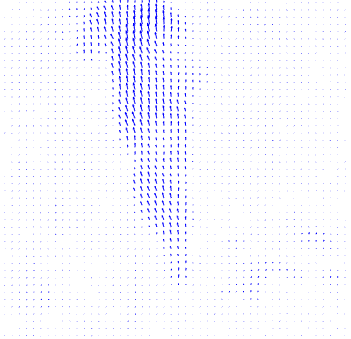}}&\fbox{\includegraphics[width=2.5cm, height=2.5cm]{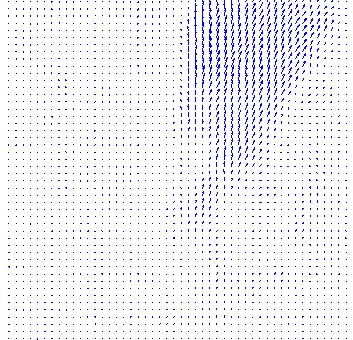}}&\fbox{\includegraphics[width=2.5cm, height=2.5cm]{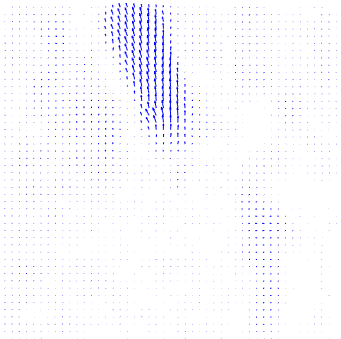}}&\fbox{\includegraphics[width=2.5cm, height=2.5cm]{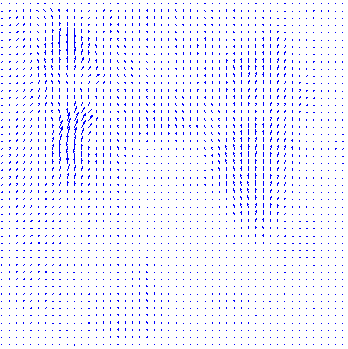}}\\
\\
\fbox{\includegraphics[width=2.5cm, height=2.5cm]{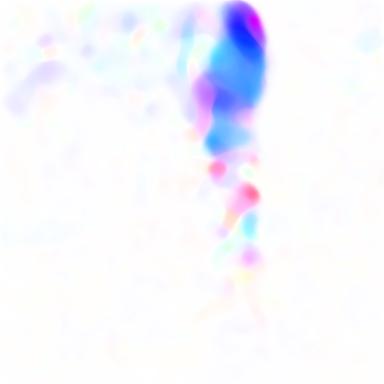}} &\fbox{\includegraphics[width=2.5cm, height=2.5cm]{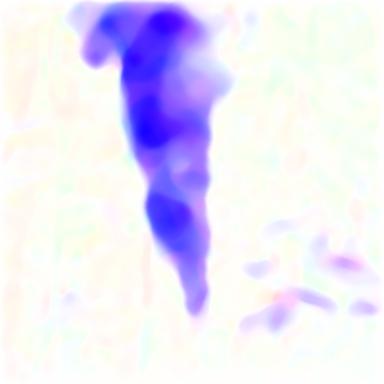}}&\fbox{\includegraphics[width=2.5cm, height=2.5cm]{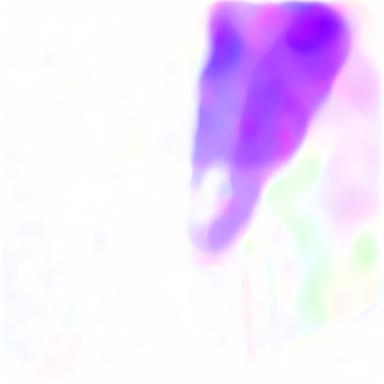}}&\fbox{\includegraphics[width=2.5cm, height=2.5cm]{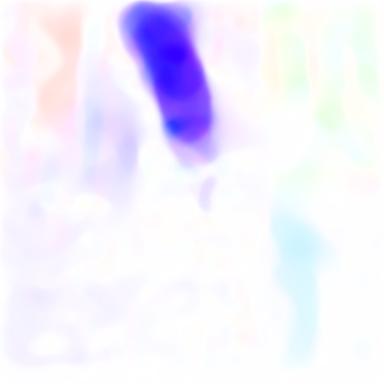}}&\fbox{\includegraphics[width=2.5cm, height=2.5cm]{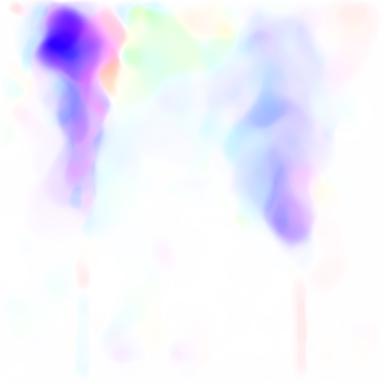}}\\
\end{tabular}
\caption{Reference images (first row), vector plots (second row), and optical flow color maps (third row) for smoke videos.}
\label{ER1}
\end{center}
\end{figure}
\begin{figure}[h!]
\begin{center}
\begin{tabular}{ccccc}
\textbf{Non-Smoke1}&\textbf{Non-Smoke2}&\textbf{Non-Smoke3}&\textbf{Non-Smoke4}&\textbf{Non-Smoke5}\\
\includegraphics[width=2.8cm, height=2.8cm]{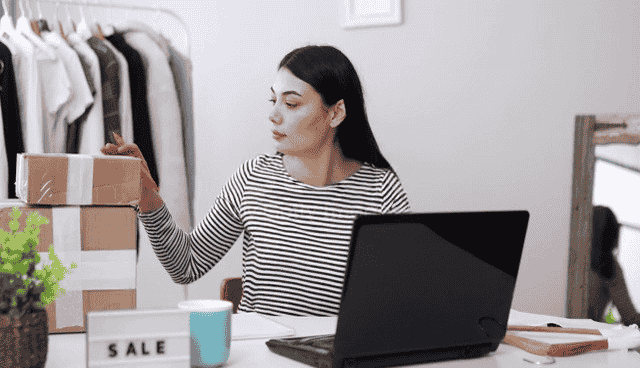}&\includegraphics[width=2.8cm, height=2.8cm]{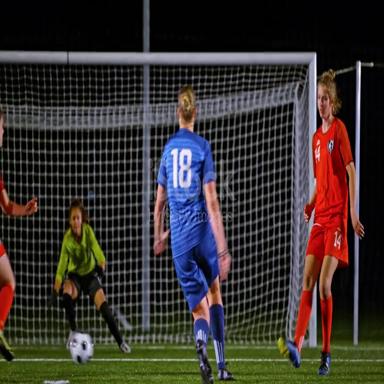}&\includegraphics[width=2.8cm, height=2.8cm]{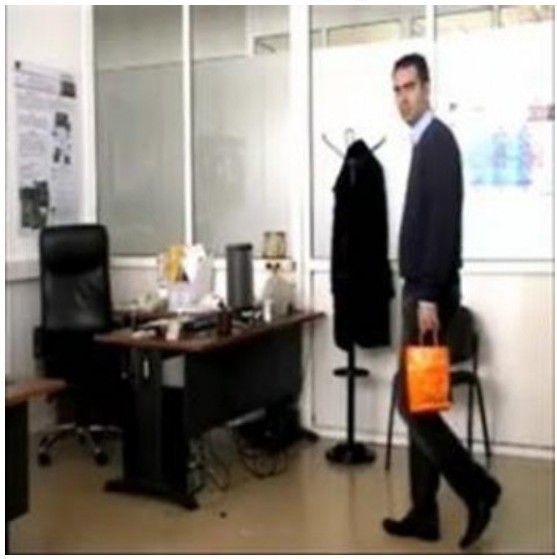}&\includegraphics[width=2.8cm, height=2.8cm]{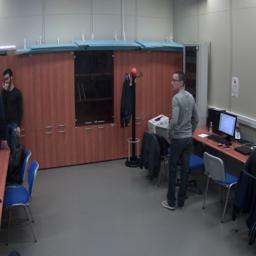}&\includegraphics[width=2.8cm, height=2.8cm]{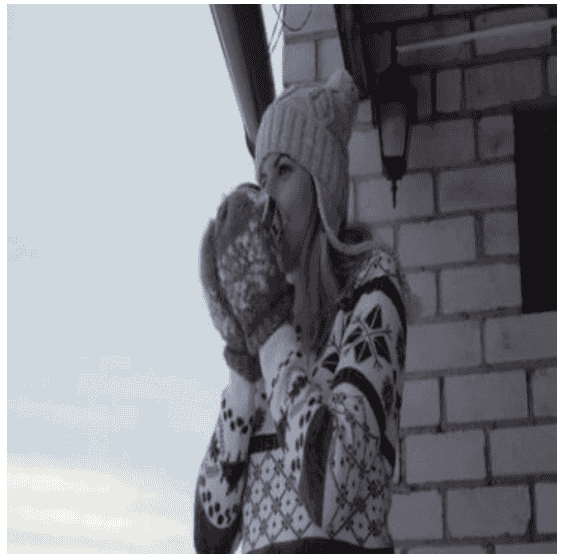}\\
\\
\fbox{\includegraphics[width=2.5cm, height=2.5cm]{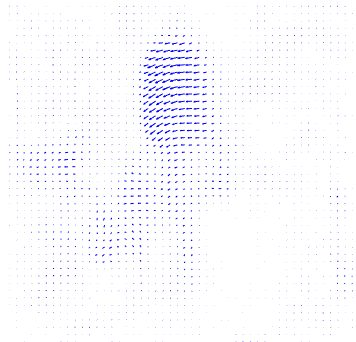}}&\fbox{\includegraphics[width=2.5cm, height=2.5cm]{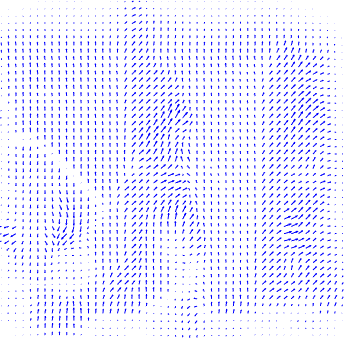}}&\fbox{\includegraphics[width=2.5cm, height=2.5cm]{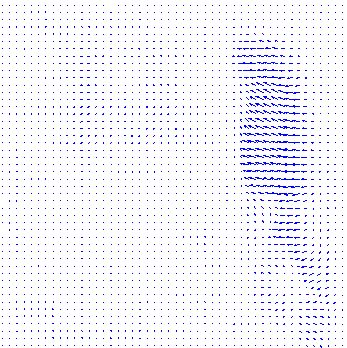}}&\fbox{\includegraphics[width=2.5cm, height=2.5cm]{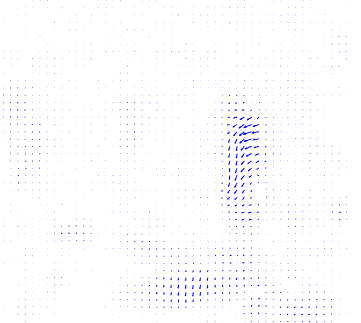}}&\fbox{\includegraphics[width=2.5cm, height=2.5cm]{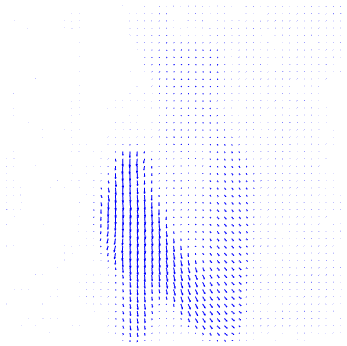}}\\
\\
\fbox{\includegraphics[width=2.5cm, height=2.5cm]{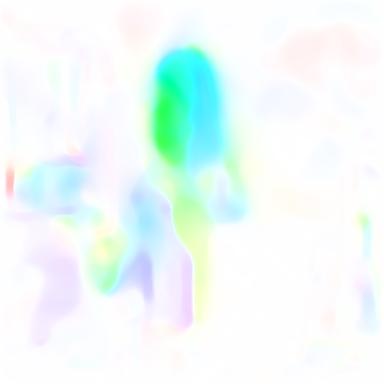}}&\fbox{\includegraphics[width=2.5cm, height=2.5cm]{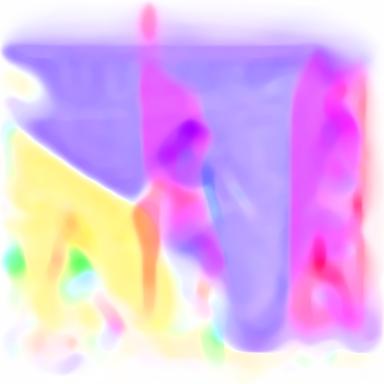}}&\fbox{\includegraphics[width=2.5cm, height=2.5cm]{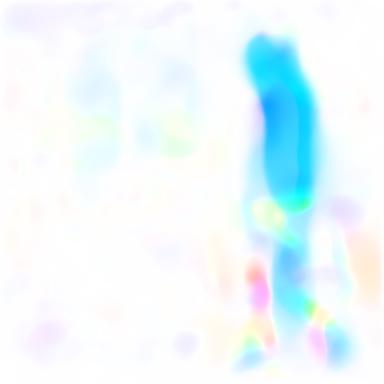}}&\fbox{\includegraphics[width=2.5cm, height=2.5cm]{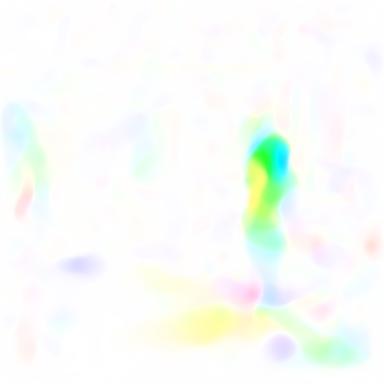}}&\fbox{\includegraphics[width=2.5cm, height=2.5cm]{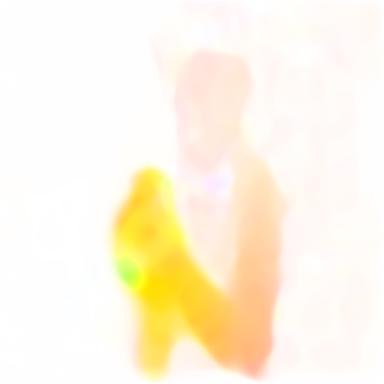}}\\
\end{tabular}
\caption{Reference images (first row), vector plots (second row), and optical flow color maps (third row) for non-smoke videos.}
\label{ER2}
\end{center}
\end{figure} 
\par
\begin{figure}[h!]
\begin{center}
\begin{tabular}{ccc}
\includegraphics[width=2.5cm, height=2.5cm]{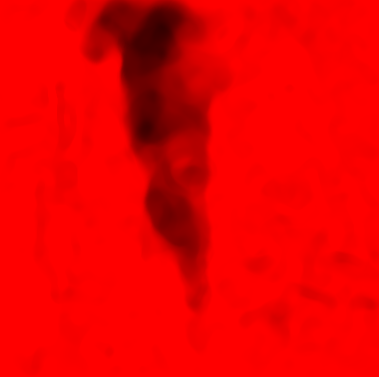}&\includegraphics[width=2.5cm, height=2.5cm]{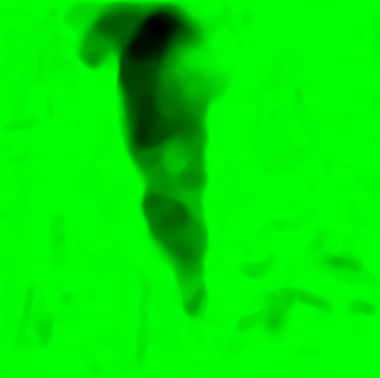}&\includegraphics[width=2.5cm, height=2.5cm]{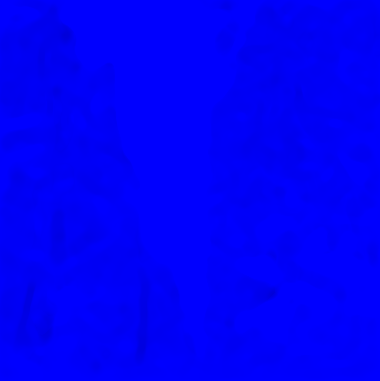}\\
\\
\includegraphics[width=2.5cm, height=2.5cm]{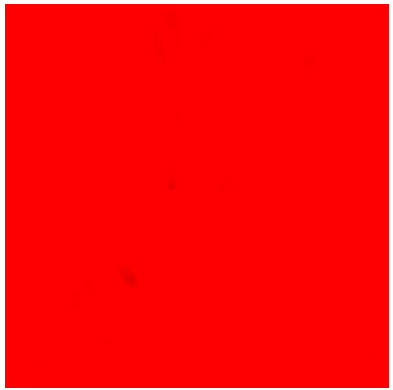}&\includegraphics[width=2.5cm, height=2.5cm]{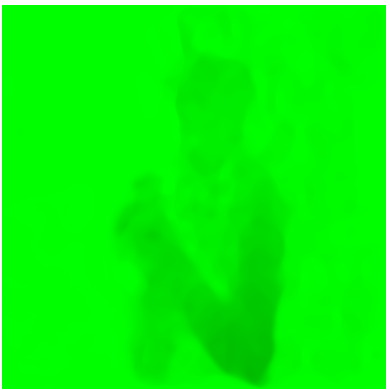}&\includegraphics[width=2.5cm, height=2.5cm]{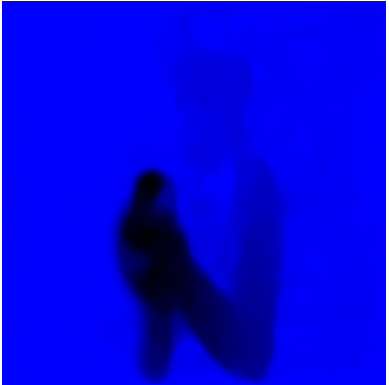}\\
\end{tabular}
\caption{First, second, and third columns correspond to the red, green, and blue channels of the smoke and non-smoke images, respectively.}
\label{OFBC}
\end{center}
\end{figure}

 The first experiment presents the optical flow estimation using the FCDLe-FOV model with parameters $\alpha=0.5, \lambda=225, \theta=0.001,$ and $\nu=1000$, and $100$ iterations. The datasets used in the experiment include scenes like forests, roads, industrial areas, and crowds, captured under both stationary and non-stationary illumination conditions. The experimental results for smoke and non-smoke images are shown in Figs.~\ref{ER1} and \ref{ER2}, with vector plots and color maps illustrated in the second and third rows, respectively. According to the concept introduced by Muller et al.~\cite{Mueller2013}, smoke predominantly moves upward, with the blue channel of the color maps showing higher sensitivity to this motion, as shown in Fig.~\ref{OFBC}. In the non-smoke class, objects often do not exhibit upward motion patterns, as seen in the blue channel of Fig.~\ref{OFBC}. Moreover, the optical flow maps effectively preserve motion boundaries, supporting the use of dual-phase level set segmentation combined with fractional-order derivatives. Furthermore, dense optical flow fields maintain motion edges, and smoke motion regions are clearly visible, confirming the reliability of the FCDLe-FOV model. 
\par
\begin{figure}[h!]
  \centering
  \includegraphics[height=6.4cm,width=15.4cm]{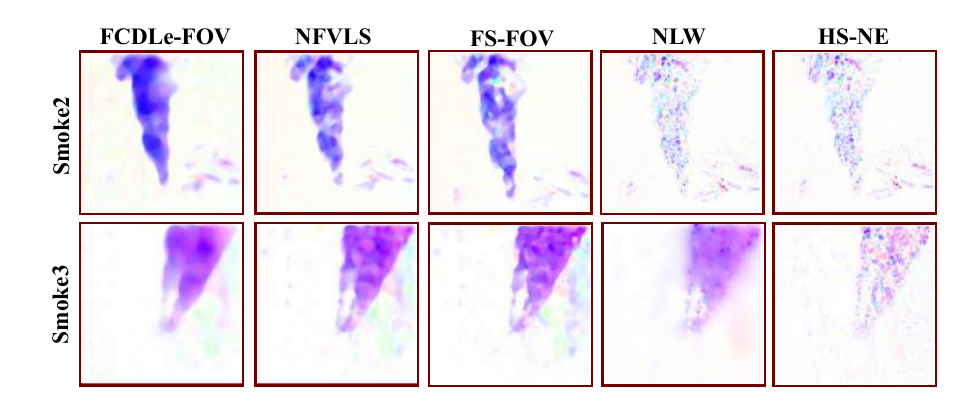}
  \caption{Qualitative comparison of FCDLe-FOV model and SOTA models on smoke datasets.}\label{OFCS}
\end{figure}

\begin{figure}[h!]
  \centering
  \includegraphics[height=6.4cm,width=15.4cm]{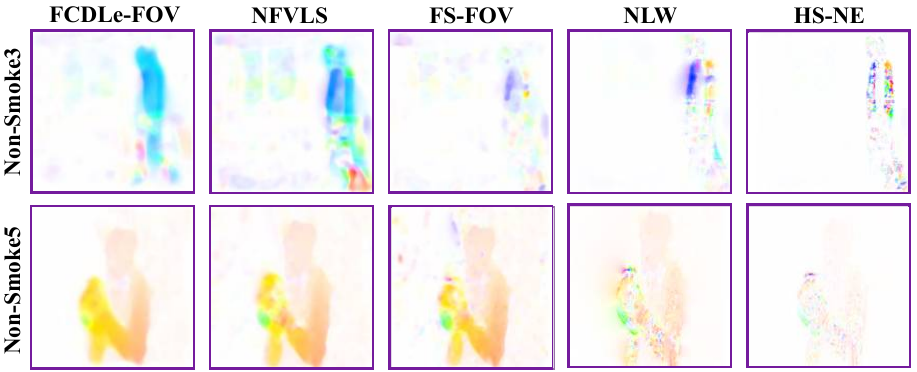}
  \caption{Qualitative comparison of FCDLe-FOV model and SOTA models on non-smoke datasets.}\label{OFCNS}
\end{figure}

 The second experiment consists of two stages to validate the effectiveness of the FCDLe-FOV model. In the initial stage, a quantitative evaluation is conducted using two Middlebury datasets and one Sintel dataset, selected for their inclusion of ground truth data, which is lacking in real-world smoke and non-smoke datasets. The model's accuracy is assessed using AAE, AEPE, and AENG metrics and compared against existing models such as NFVLS~\cite{Muzammil2023}, FS-FOV~\cite{Lu2019a}, NLW~\cite{Huang2020}, and HS-NE~\cite{Kumar2016a}. As shown in Table~\ref{CT}, FCDLe-FOV model outperforms all compared models across these metrics. The second stage presents a qualitative comparison, as illustrated in Figs.~\ref{OFCS} and \ref{OFCNS}, which show optical flow color maps for two smoke scenes (Smoke2 and Smoke3) and two non-smoke scenes (Non-Smoke3 and Non-Smoke5), respectively. Together, these visual representations of color maps demonstrate that the FCDLe-FOV model generates more accurate results and preserves motion boundaries more effectively than other models. Therefore, the results validate the effectiveness of the FCDLe-FOV model in handling complex smoke and non-smoke scenarios.
\begin{table*}[h!]
\centering
\caption{Performance comparison of FCDLe-FOV model with other optical flow models.}
\footnotesize{
\resizebox{\textwidth}{!}{%
\begin{tabular}{p{3.3cm}ccccccccc}
\toprule \midrule
{\textbf{\cellcolor{blue!10}Datasets}} & \multicolumn{3}{c}{\textbf{\cellcolor{blue!10}Middlebury1}} & \multicolumn{3}{c}{\textbf{\cellcolor{blue!10}Middlebury2}} & \multicolumn{3}{c}{\textbf{\cellcolor{blue!10}Sintel1}}  \\ \midrule \midrule
    \textbf{Models}                       & \textbf{AAE} & \textbf{AEPE}  & \textbf{AENG} & \textbf{AAE} & \textbf{AEPE}  & \textbf{AENG} & \textbf{AAE} & \textbf{AEPE}  & \textbf{AENG}\\ \cmidrule(lr){1-1}  \cmidrule(lr){2-4} \cmidrule(lr){5-7} \cmidrule(lr){8-10}         
FCDLe-FOV                      & 0.173 & 0.877  & 1.875 & 0.176 & 2.330  &  2.461 & 0.101 & 0.393  & 1.079   \\
NFVLS~\cite{Muzammil2023}                   & 0.211 & 1.159  & 3.397 & 0.231 & 3.669  & 5.363& 0.114 & 0.516  & 1.783 \\
FS-FOV~\cite{Lu2019a}          & 0.212 & 0.997  & 2.762 & 0.242& 3.407  & 4.784 & 0.126 & 0.542  & 1.803 \\
NLW~\cite{Huang2020}          & 0.378 & 2.248  & 5.329 & 0.605 & 5.812 & 8.471 & 0.274 & 1.184 & 3.919  \\
HS-NE~\cite{Kumar2016a} & 0.782 & 3.051  & 6.835 & 0.451 & 2.023  & 3.488 & 0.451 & 2.023  & 3.488  \\ \bottomrule
\end{tabular}}
}
\label{CT}
\end{table*}
\par The primary objective of the third experiment is to evaluate the robustness of the FCDLe-FOV model in the presence of Salt-and-pepper, Gaussian, and Poisson noise. These noise types are randomly introduced into four image sequences, comprising two smoke scenarios (Smoke1 and Smoke4) and two non-smoke scenarios (Non-Smoke1 and Non-Smoke4). Both Salt-and-pepper noise and Gaussian noise are added with a mean of zero and a standard deviation of $0.01$, while Poisson noise is inherently present in the datasets rather than artificially introduced. The robustness of the FCDLe-FOV model is assessed using the SSIM score, with results in Table~\ref{RS} demonstrating that it maintains high SSIM values, confirming its effectiveness in handling various noise types while preserving structural details in optical flow color maps.
\begin{table}[hb!]
\caption{SSIM scores of the FCDLe-FOV model on smoke and non-smoke images under different noise conditions.}
  \centering
  \begin{tabular}{lcccc}
    \toprule \toprule
    {\cellcolor{blue!10}\textbf{Datasets}}&\multicolumn{2}{c} {\cellcolor{blue!10}{\textbf{Smoke Datasets}}}&\multicolumn{2}{c}{\cellcolor{blue!10}{\textbf{Non-Smoke Datasets}}}\\ \midrule \midrule
    \textbf{Noises} & \textbf{Smoke1} & \textbf{Smoke4} & \textbf{Non-Smoke1} & \textbf{Non-Smoke4}\\ \cmidrule(lr){1-1}  \cmidrule(lr){2-3} \cmidrule(lr){4-5}       
    \textbf{Gaussian noise} & 0.87 & 0.91 & 0.97 & 0.95\\
    \textbf{Poisson noise} & 0.95 &  0.97 & 0.98 & 0.98\\
    \textbf{Salt-and-pepper noise} & 0.97 &  0.98 & 0.99 & 0.99\\
    \bottomrule
  \end{tabular}
  \label{RS}
\end{table}
\begin{figure}[b!]
\begin{center}
\begin{tabular}{ccccc}
\includegraphics[width=2.9cm, height=2.9cm]{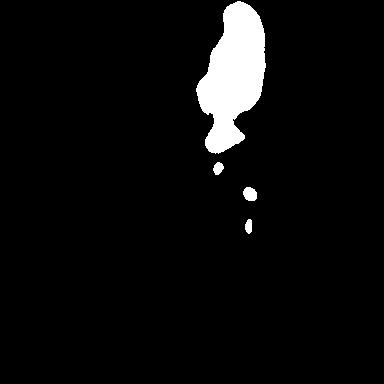} &\includegraphics[width=2.9cm, height=2.9cm]{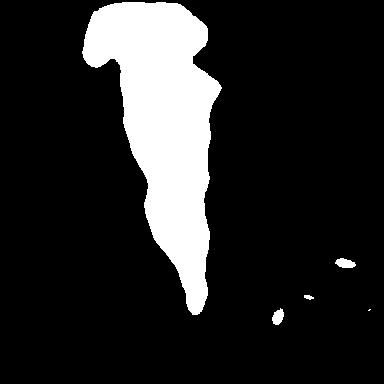}&\includegraphics[width=2.9cm, height=2.9cm]{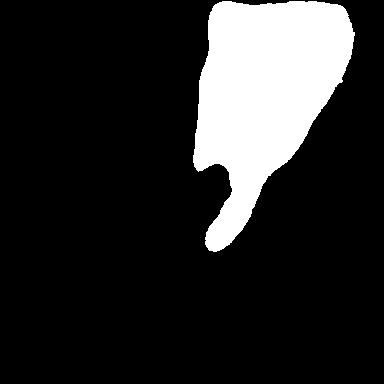}&\includegraphics[width=2.9cm, height=2.9cm]{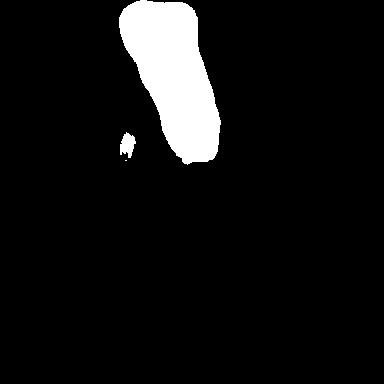}&\includegraphics[width=2.9cm, height=2.9cm]{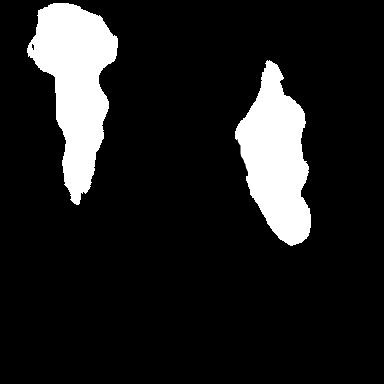}\\
\\
\includegraphics[width=2.9cm, height=2.9cm]{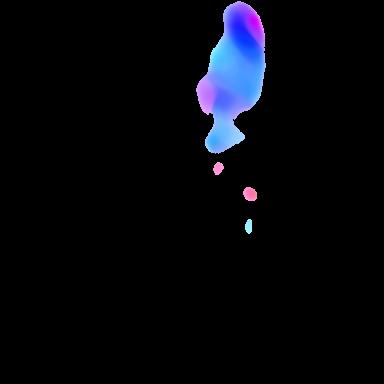} &\includegraphics[width=2.9cm, height=2.9cm]{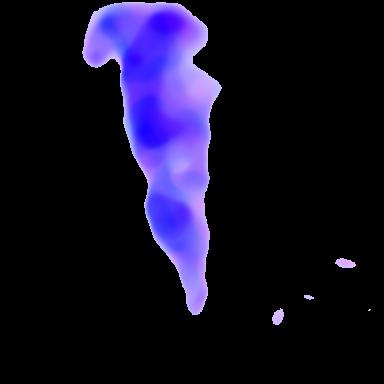}&\includegraphics[width=2.9cm, height=2.9cm]{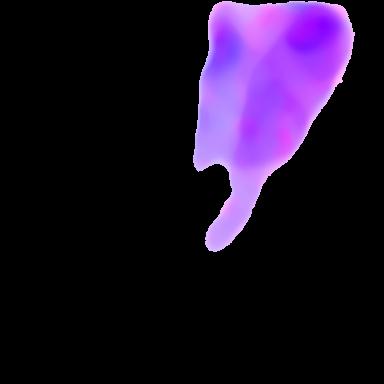}&\includegraphics[width=2.9cm, height=2.9cm]{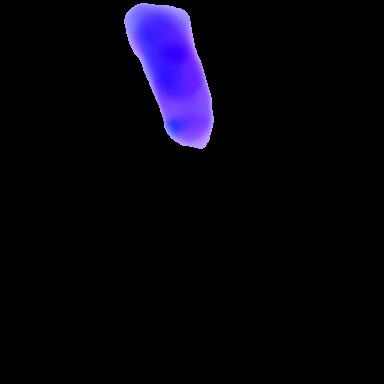}&\includegraphics[width=2.9cm, height=2.9cm]{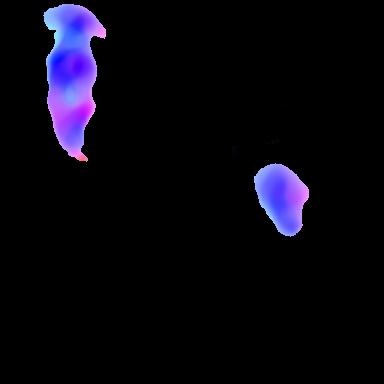}\\
\\
\includegraphics[width=2.9cm, height=2.9cm]{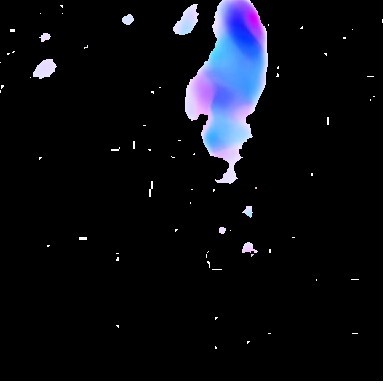} &\includegraphics[width=2.9cm, height=2.9cm]{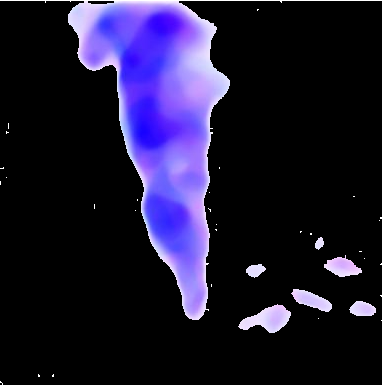}&\includegraphics[width=2.9cm, height=2.9cm]{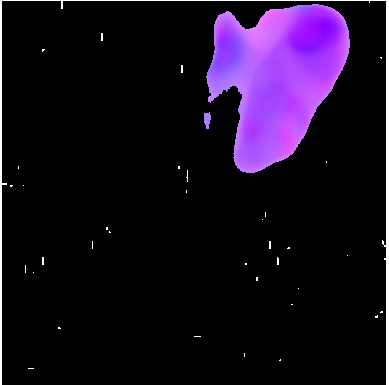}&\includegraphics[width=2.9cm, height=2.9cm]{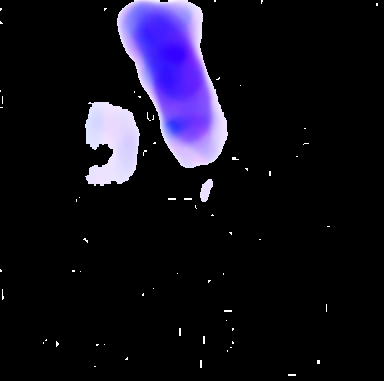}&\includegraphics[width=2.9cm, height=2.9cm]{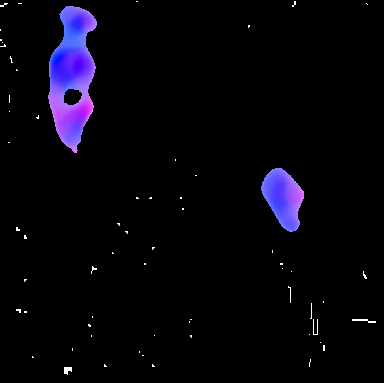}\\
\\
\includegraphics[width=2.9cm, height=2.9cm]{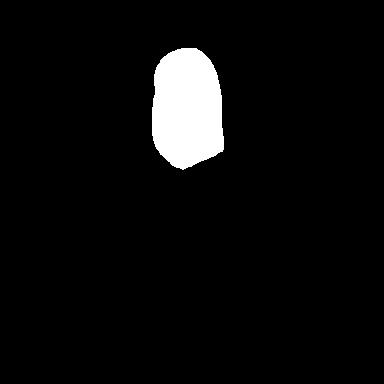}&\includegraphics[width=2.9cm, height=2.9cm]{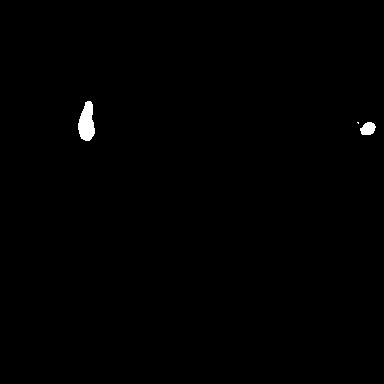}&\includegraphics[width=2.9cm, height=2.9cm]{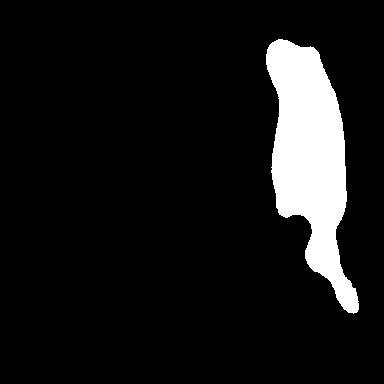}&\includegraphics[width=2.9cm, height=2.9cm]{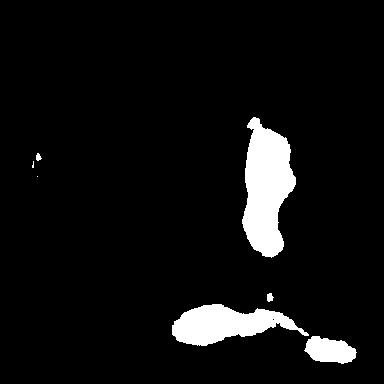}&\includegraphics[width=2.9cm, height=2.9cm]{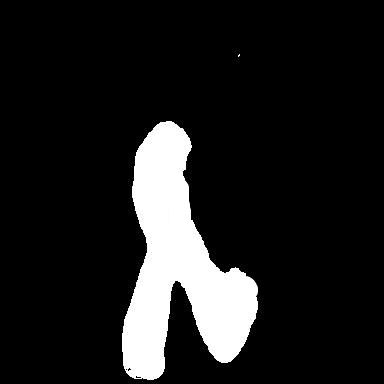}\\
\\
\includegraphics[width=2.9cm, height=2.9cm]{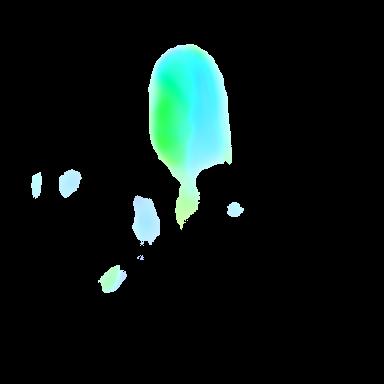}&\includegraphics[width=2.9cm, height=2.9cm]{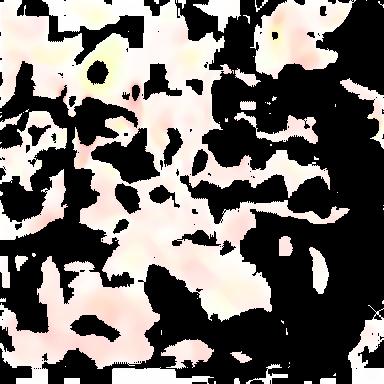}&\includegraphics[width=2.9cm, height=2.9cm]{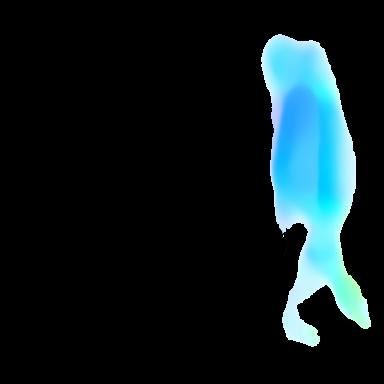}&\includegraphics[width=2.9cm, height=2.9cm]{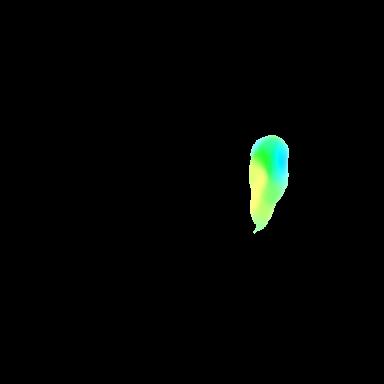}&\includegraphics[width=2.9cm, height=2.9cm]{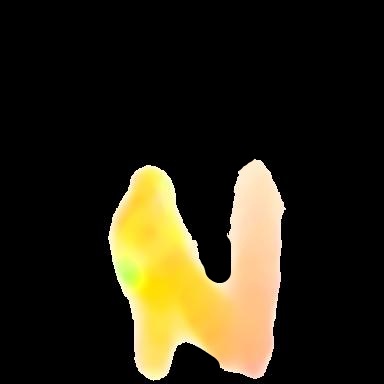}\\
\\
\includegraphics[width=2.9cm, height=2.9cm]{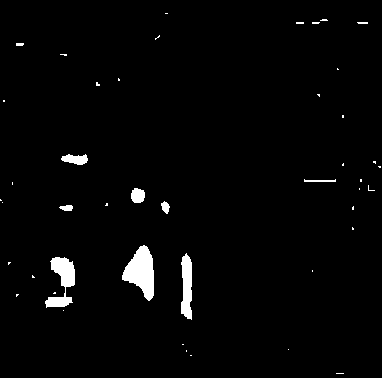}&\includegraphics[width=2.9cm, height=2.9cm]{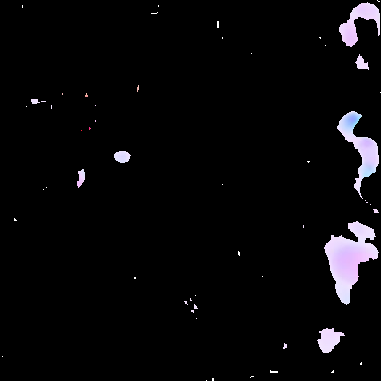}&\includegraphics[width=2.9cm, height=2.9cm]{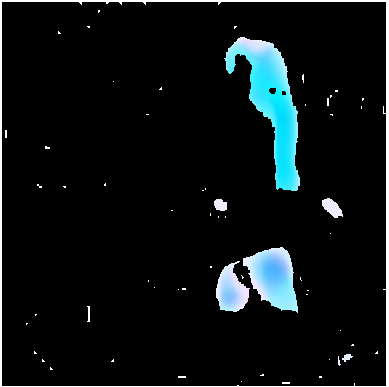}&\includegraphics[width=2.9cm, height=2.9cm]{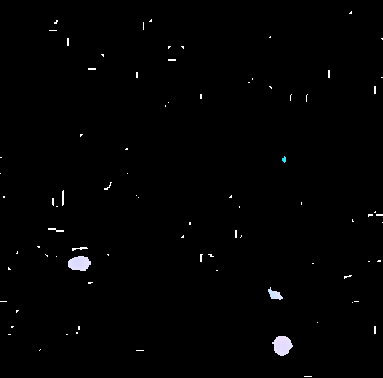}&\includegraphics[width=2.9cm, height=2.9cm]{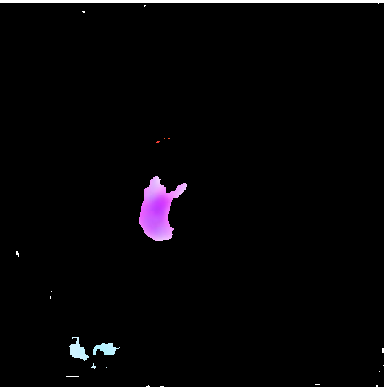}\\
\end{tabular}
\caption{Binary mask (first and fourth row), segmented color map (second and fifth row), and comparison of GMM technique with existing method~\cite{Khan2023} (third and sixth row) for smoke and non-smoke images.}
\label{SBM}
\end{center}
\end{figure}
\par The fourth experiment aims to effectively highlight smoke-active regions by generating a binary mask that suppresses background features such as vehicles, trees, the sky, and human motion. This binary mask is derived from the estimated color maps using the GMM technique, which effectively distinguishes between smoke and non-smoke regions based on their statistical distributions. Subsequently, the resulting binary mask is multiplied by the optical flow color maps to produce segmented color maps. In Fig.~\ref{SBM}, the first and fourth rows depict the binary masks of the smoke and non-smoke images, respectively, while the second and fifth rows show the segmented optical flow color maps for the corresponding smoke and non-smoke image frames. These are based on the reference images in Figs.~\ref{ER1} and~\ref{ER2} for the smoke and non-smoke images, respectively. The second and third rows of this figure compare the results of the GMM technique with the method proposed by Khan et al.~\cite{Khan2023} for smoke segmentation, while the fifth and sixth rows provide the corresponding comparison for non-smoke segmentation. This comparison reveals that the proposed binary masks using GMM technique accurately segment the smoke regions by capturing their distinct intensity and texture characteristics. In the case of the non-smoke class, the model often segments background objects separately, as reflected in the binary masks shown in the fourth row of Fig.~\ref{SBM}. However, the segmented optical flow color maps indicate that interference from background motion is highly suppressed. This confirms the effectiveness of the proposed GMM technique in isolating the smoke-related motion while minimizing interference from background movements and leading to more precise segmentation of smoke regions. 
\enlargethispage{35pt}
\par Optical flow-based motion information captures the dynamic and complex behavior of smoke, while reference images provide complementary spatial information. By fusing segmented optical flow maps with reference images, the system effectively leverages both motion and appearance cues to enhance the detection accuracy of the proposed TP-UAST model.
\par
\begin{figure}[h!]
\centering
\begin{tabular}{ccc}
\includegraphics[height=3.5cm,width=5cm]{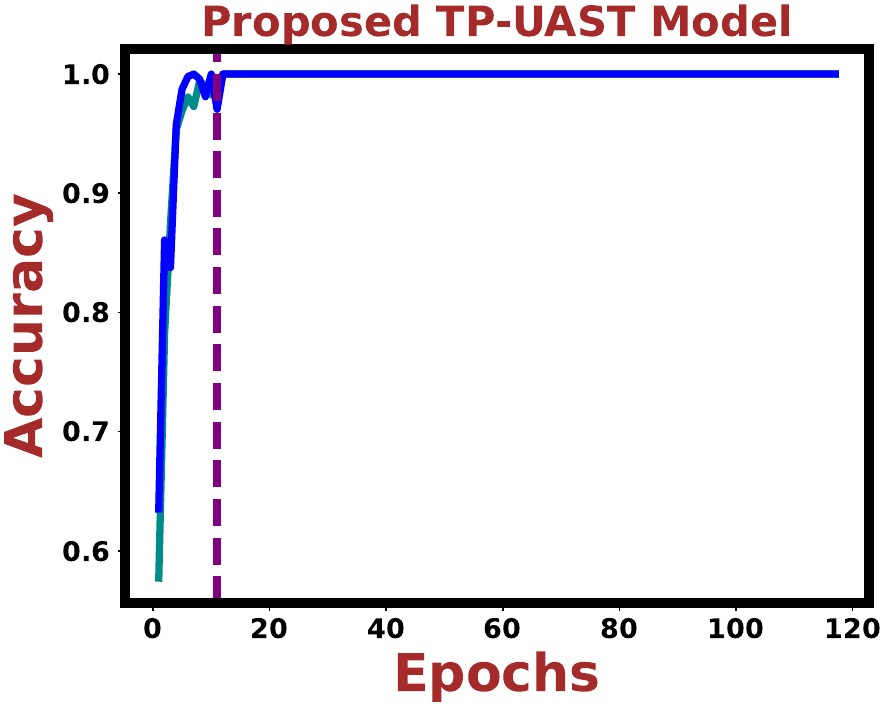}&\includegraphics[height=3.5cm,width=5cm]{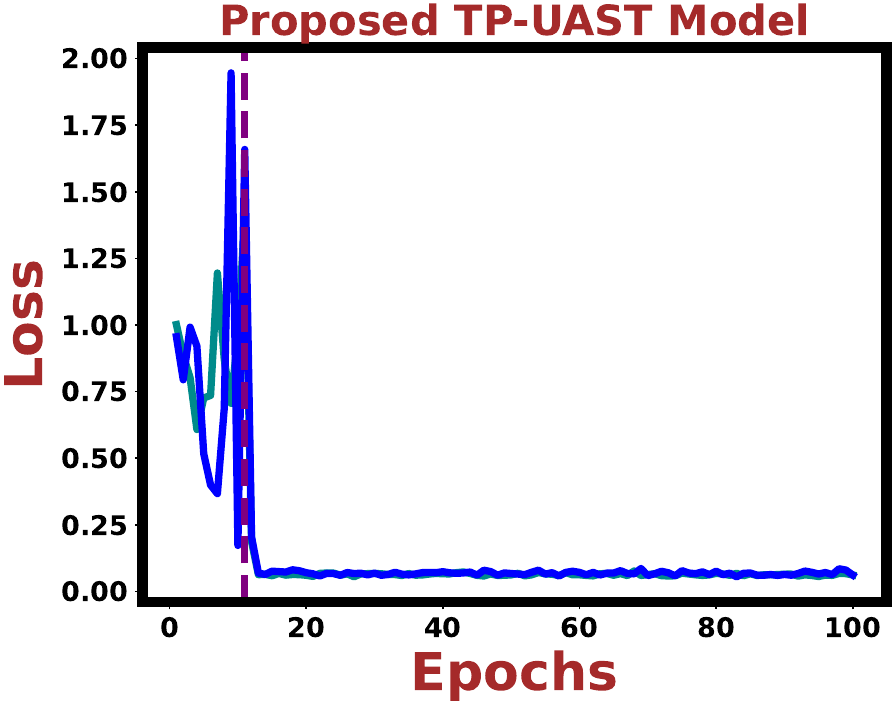}&\includegraphics[height=3.5cm,width=5cm]{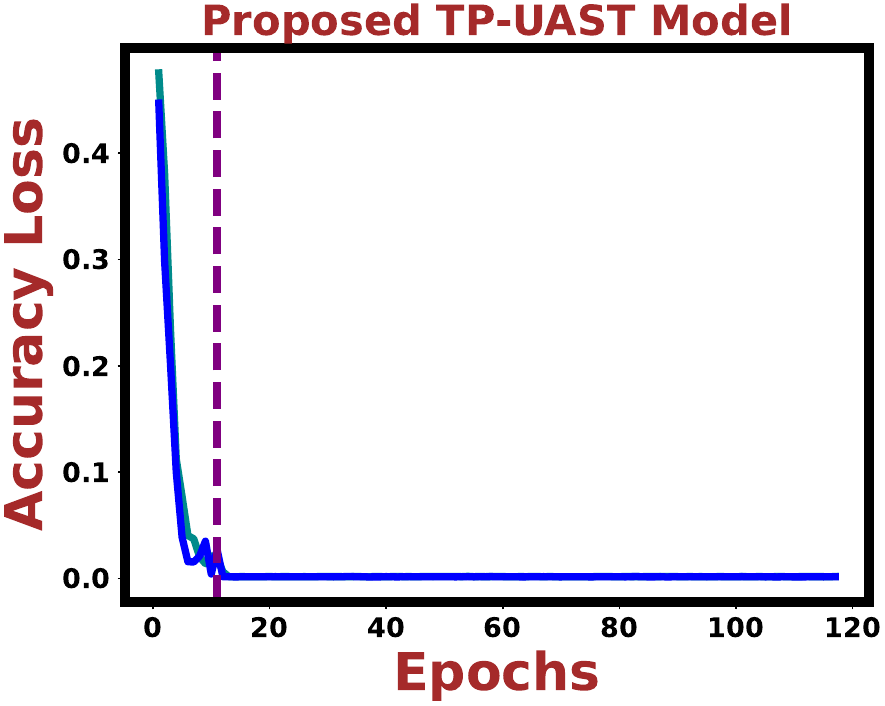}\\
\mbox{\ \ \ \ \ \ }\textbf{(a)}&\mbox{\ \ \ \ \ \ \ }\textbf{(b)}&\mbox{\ \ \ \ \ \ \ }\textbf{(c)}\\
\end{tabular}
\caption{Learning curves in terms of (a) accuracy, (b) loss, and (c) loss in accuracy.}\label{LC}
\end{figure}

 The fifth experiment analyzes the learning curves of the TP-UAST model in terms of accuracy, loss, and loss in accuracy across two phases, as shown in Fig.~\ref{LC}. The vertical dashed line at epoch $11$ marks the point of convergence, after which changes are minimal. In Fig.~\ref{LC}(a), during Phase I, the accuracy curve rises rapidly toward $1.0$, indicating the rapid convergence of the TP-UAST model. In phase II, the curve remains stable. Similar convergence behavior is demonstrated by the loss and loss in accuracy curves shown in Fig.~\ref{LC}(b) and Fig.~\ref{LC}(c), respectively. Thus, these curves demonstrate the TP-UAST's efficiency and stability. 
\par
\begin{figure}[]
\centering
\begin{tabular}{cc}
\includegraphics[height=5.9cm,width=7.5cm]{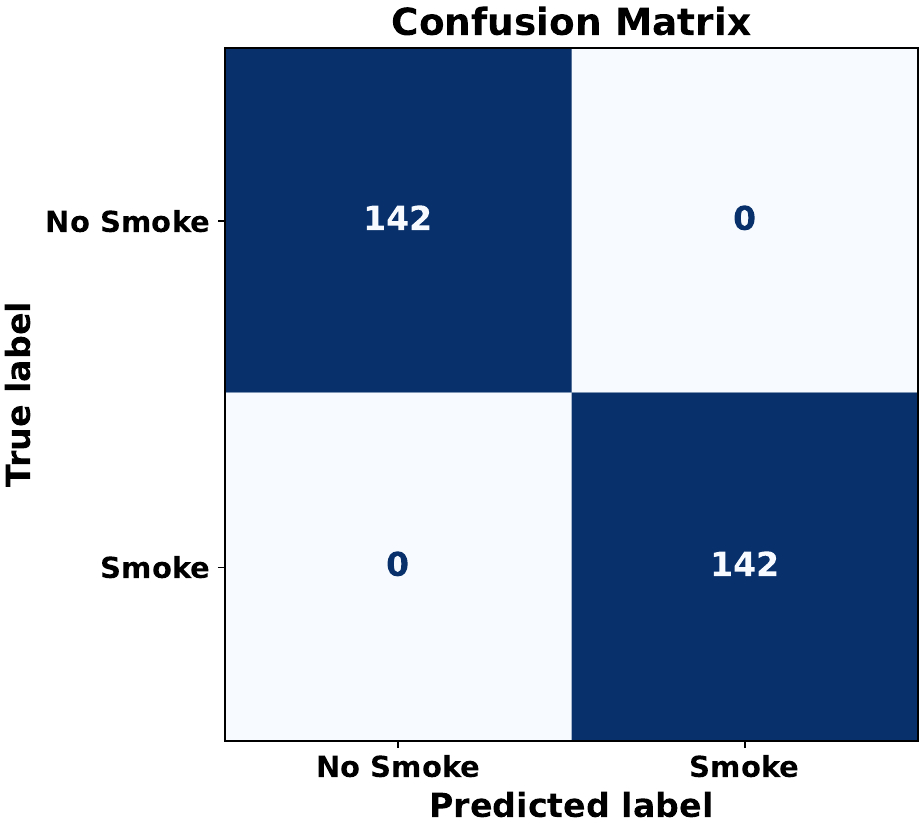}&\includegraphics[height=5.9cm,width=7.5cm]{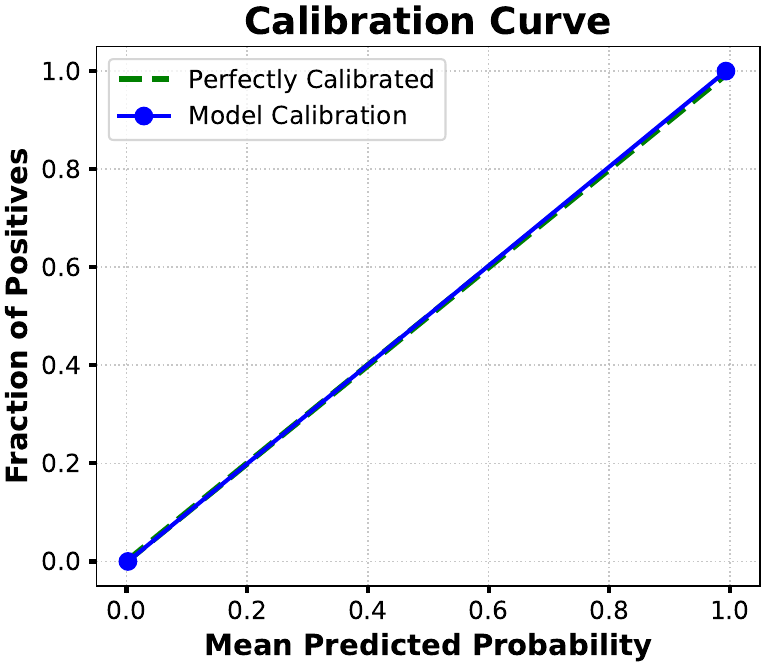}\\
\mbox{\ \ \ \ \ \ \ \ \ \ }\textbf{(a)}&\mbox{\ \ \ \ \ \ }\textbf{(b)}\\
\end{tabular}
\caption{Demonstrating (a) Confusion matrix and (b) reliability diagram for TP-UAST model.}\label{CMC}
\end{figure}

 The sixth experiment presents Table~\ref{Table_1} for a comparison of TP-UAST against leading smoke-detection algorithms. Remarkably, TP-UAST achieves a satisfactory score of $1.00$ across accuracy, precision, recall and F{1}-score, yielding a balanced classifier with zero false positives and false negatives on the validation set. By contrast, the best competing method MFFNet~\cite{Liu2025a} attains 0.98 accuracy and 0.94 precision, while Safarov et al.~\cite{Safarov2024} achieve 0.96 precision and 0.97 recall. Yang et al.~\cite{Yang2023} and the energy-efficient model~\cite{Almeida2022} both report balanced performance metrics in the range of 0.94-0.95. Moreover, YOLOv2~\cite{Saponara2021} attains 0.96 accuracy and balances precision against recall, while classic deep networks such as ResNet50~\cite{He2016} and VGG16~\cite{Simonyan2014} suffer from trade-offs between recall and precision. Other models like TFNet~\cite{Wang2025}, YOLOv8s~\cite{Kong2024}, and HPO-YOLOv5~\cite{Sozol2025} perform moderately. Methods such as DBN~\cite{Pundir2017} and the lightweight architecture of MobileNet~\cite{Howard2017} further underperform in either precision or recall. These results underscore the efficacy of the TP-UAST model and the fused optical-flow and appearance representation. Additionally, TP-UAST's discriminatory performance is also illustrated by the confusion matrix as shown in Fig.~\ref{CMC}(a).   
\begin{table}[t!]
\begin{center}
\caption{A comparison of proposed TP-UAST model against SOTA models.\label{Table_1}}
{\footnotesize
\resizebox{\textwidth}{!}{%
 \begin{tabular}{p{6cm}cccc}
 \toprule \midrule
 \textbf{Algorithm} & \textbf{Accuracy} & \textbf{Precision} & \textbf{Recall} & \textbf{F1-Score}\\ \midrule \midrule 
 \textbf{TP-UAST model} & \boldmath{$1.00$} & \boldmath{$1.00$} & \boldmath{$1.00$} & \boldmath{$1.00$}\\  
 HPO-YOLOv5~\cite{Sozol2025} & $-$ & $0.93$ &$0.86$  & $0.89$\\  
 MFFNet~\cite{Liu2025a} & $0.98$ & $0.94$ &$-$ $-$ & $-$ $-$ \\  
 TFNet~\cite{Wang2025} & $-$ $-$& $0.82$ &$0.75$& $0.78$ \\  
 YOLOv8s~\cite{Kong2024} & $-$ $-$ & $0.91$ & $0.85$ & $0.88$ \\  
 Safarov et al.~\cite{Safarov2024}& $-$ $-$ & $0.96$ & $0.97$ & $-$ $-$\\  
 Yang et al.~\cite{Yang2023}& $-$ $-$ & $0.94$ & $0.93$ & $-$ $-$\\  
 Energy efficient model~\cite{Almeida2022} & $0.95$ & $0.95$ & $0.95$ & $0.95$\\  
 DLBN~\cite{Wu2021} & $0.89$ & $0.98$ & $0.81$ & $0.89$ \\  
 YOLOv2~\cite{Saponara2021} & $0.96$ & $0.97$ & $0.95$ & $0.97$ \\  
 DBN~\cite{Pundir2017} & $0.94$ & $0.93$ & $0.96$ & $0.94$ \\  
 MobileNet model~\cite{Howard2017} & $0.88$ & $0.81$ & $0.99$ & $0.89$\\  
 ResNet50 model~\cite{He2016} & $0.89$ & $0.81$ & $1.00$ & $0.89$\\  
 ResNet101 model~\cite{He2016} & $0.95$ & $0.95$ & $0.93$ & $0.94$\\  
 GoogleNet model~\cite{Ballester2016} & $0.89$ & $0.85$ & $0.97$ & $0.90$\\  
 VGG16 model~\cite{Simonyan2014} & $0.81$ & $0.72$ & $1.00$ & $0.84$\\ \bottomrule
 \end{tabular}
 }
 }
 \end{center}
 \end{table}
\par The seventh experiment is described by the Fig.~\ref{CMC}(b), which shows the TP-UAST model's reliability diagram on the test set. Here, the mean of the predicted probabilities, denoted by blue markers, are plotted against the empirical fraction of positives. The number of bins taken is ten. The calibration curve lies essentially very close to the ideal diagonal (green dashed line) over the entire \([0,1]\) range, yielding an ECE approximately equal to $0$.  This near-perfect alignment demonstrates that TP-UAST's predicted probability estimates faithfully reflect true outcome frequencies. Such strong calibration is critical, since it allows end-users to set actionable probability thresholds with known reliability, which is $0.5$ in the proposed study.
\par
\begin{figure}[]
\centering
\begin{tabular}{cc}
\includegraphics[height=5.3cm,width=7.8cm]{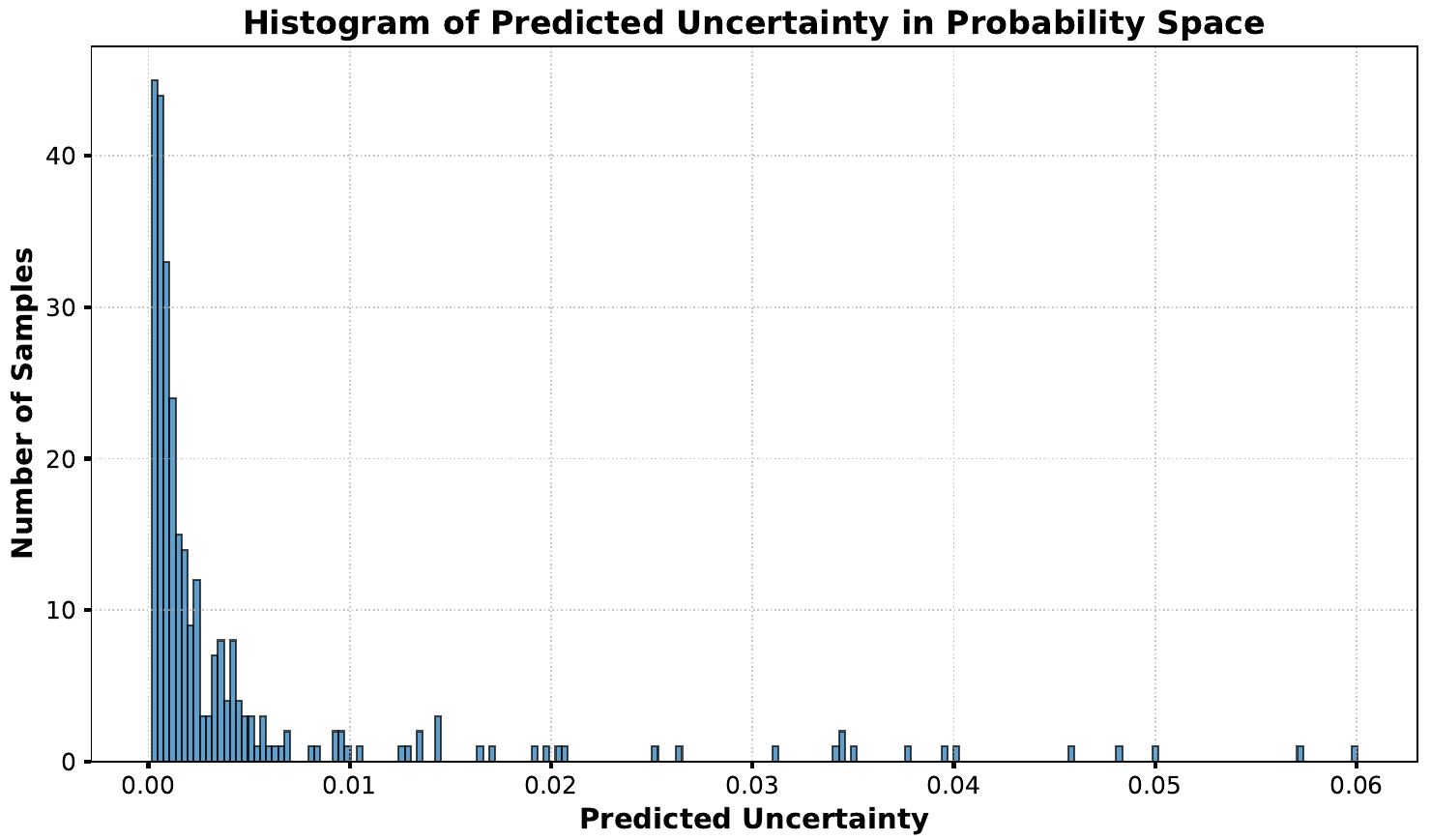} & \includegraphics[height=5.3cm,width=7.8cm]{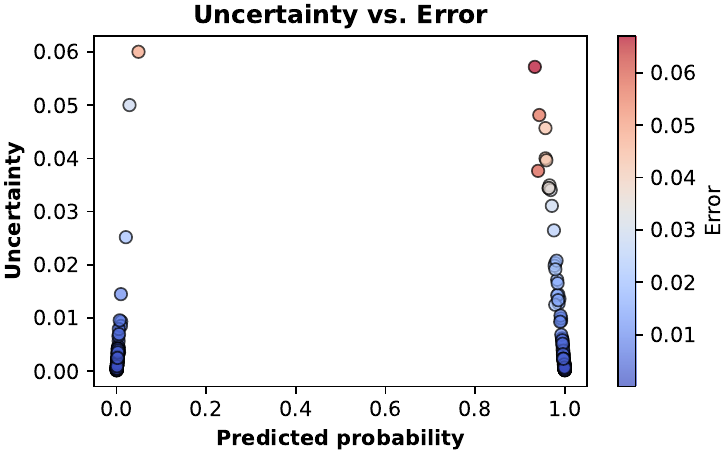} \\
\textbf{(a)}&\textbf{(b)}\\
\multicolumn{2}{c}{\includegraphics[height=5.8cm,width=8.3cm]{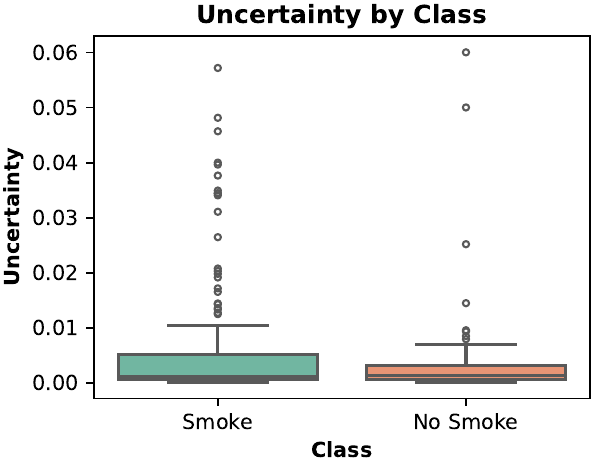}} \\
\multicolumn{2}{c}{\mbox{\ \ \ \ \ \ \ \ }\textbf{(c)}}\\
\end{tabular}
\caption{(a) Histogram of predicted uncertainty in probability space, (b) uncertainty vs. error plot, and (c) uncertainty by class for TP-UAST model.}\label{UP}
\end{figure}

 In the eighth experiment, Fig.~\ref{UP}(a) presents the distribution of per-sample predictive uncertainty $\sigma_{p}$ in probability space. The histogram is sharply peaked at low values $<0.005$, with a long tail extending to $\approx0.06$, indicating that most predictions are made with high confidence while a minority of cases exhibit elevated uncertainty. Fig.~\ref{UP}(b) overlays each sample's uncertainty against its absolute error \(\lvert PL - TL\rvert\). A clear positive trend emerges: points with \(\sigma_{p}>0.02\) are disproportionately associated with larger errors, whereas low-uncertainty predictions cluster near zero error. This correlation validates \(\sigma_{p}\) as an effective proxy for identifying potentially misclassified or ambiguous inputs. Fig.~\ref{UP}(c) shows class-stratified boxplots of \(\sigma_{p}\) for the "Smoke" and "No Smoke" classes. Both classes exhibit a median uncertainty around $0.002$, but the "Smoke" distribution has a heavier upper quartile and more extreme outliers, reflecting greater aleatoric variability when detecting smoke plumes. Together, these analyses demonstrate that TP-UAST's uncertainty estimates are well-calibrated, informative of error, and sensitive to class-dependent difficulty.

\par
\begin{figure}[]
  \centering
  \begin{tabular}{cc}
 \includegraphics[height=6.2cm,width=7.7cm]{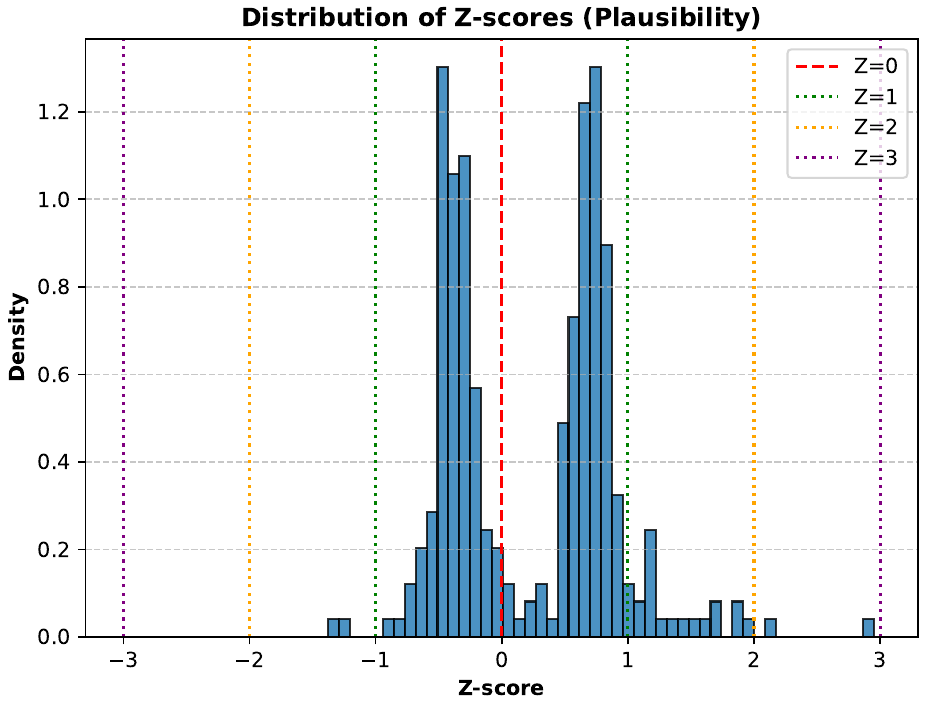} & \includegraphics[height=6.2cm,width=7.7cm]{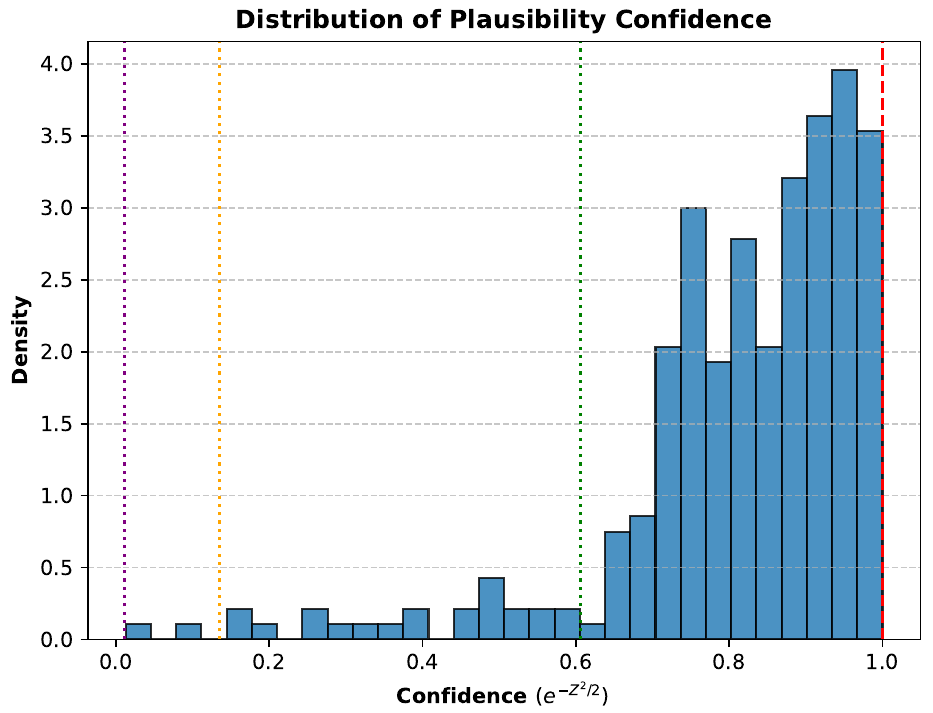}\\
    \mbox{\ \ }\textbf{(a)}&\mbox{\ \ }\textbf{(b)}\\
  \end{tabular}
  \caption{(a) Distribution of Z-scores (plausibility) and distribution of plausibility confidence for TP-UAST model.}\label{ZPG}
\end{figure}

 The ninth experiment is exhibited in the Fig.~\ref{ZPG}(a), which plots the empirical distribution of per-sample Z-scores $Z=(\ell-\mu)/\sigma_{p}$, obtained from the posterior predictive draws $\ell\sim\mathcal{N}(\mu,\sigma^{2})$. The histogram is tightly centered around \(Z=0\) with approximately $68$\% of values in $\lvert Z\rvert\le1$ and $95$\% in $\lvert Z\rvert\le2$.  Tail events beyond \(\lvert Z\rvert>2\) remain scarce, indicating that the model's predicted $\sigma_{p}$ accurately captures the scatter of its own logits. Figure~\ref{ZPG}(b) shows the corresponding distribution of plausibility confidences $ C = \exp\bigl(-\tfrac12Z^{2}\bigr)$, peaked near $C=1$ with a smooth decay toward $0.2$. Over $80$\% of samples achieve $C>0.8$, confirming that most predictions lie well within their predicted uncertainty bounds, while the low-confidence tail correctly flags rare, less-plausible draws.  Together, these results validate that TP-UAST's two-phase uncertainty learning yields self-consistent posterior predictive distributions, enabling reliable per-sample confidence estimates. Accordingly, \(C\) is partitioned into four operational tiers: High, Moderate, Low, and Very Low Confidence, facilitating threshold-based decision-making.

\par
\begin{figure}[h!]
\begin{center}
 \begin{tabular}{cccc}
 \includegraphics[width=3.5cm, height=3.5cm]{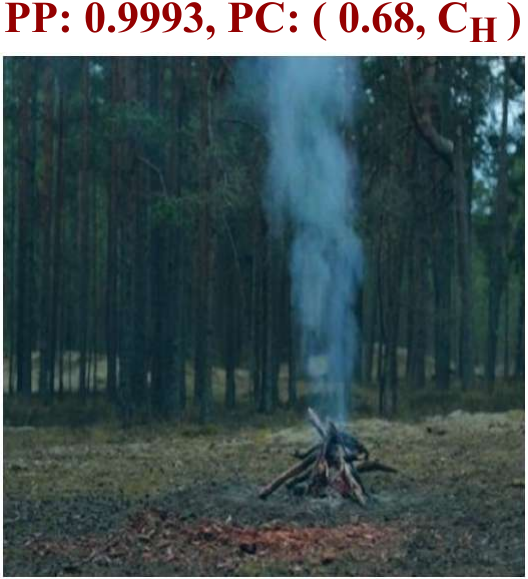} & \includegraphics[width=3.5cm, height=3.5cm]{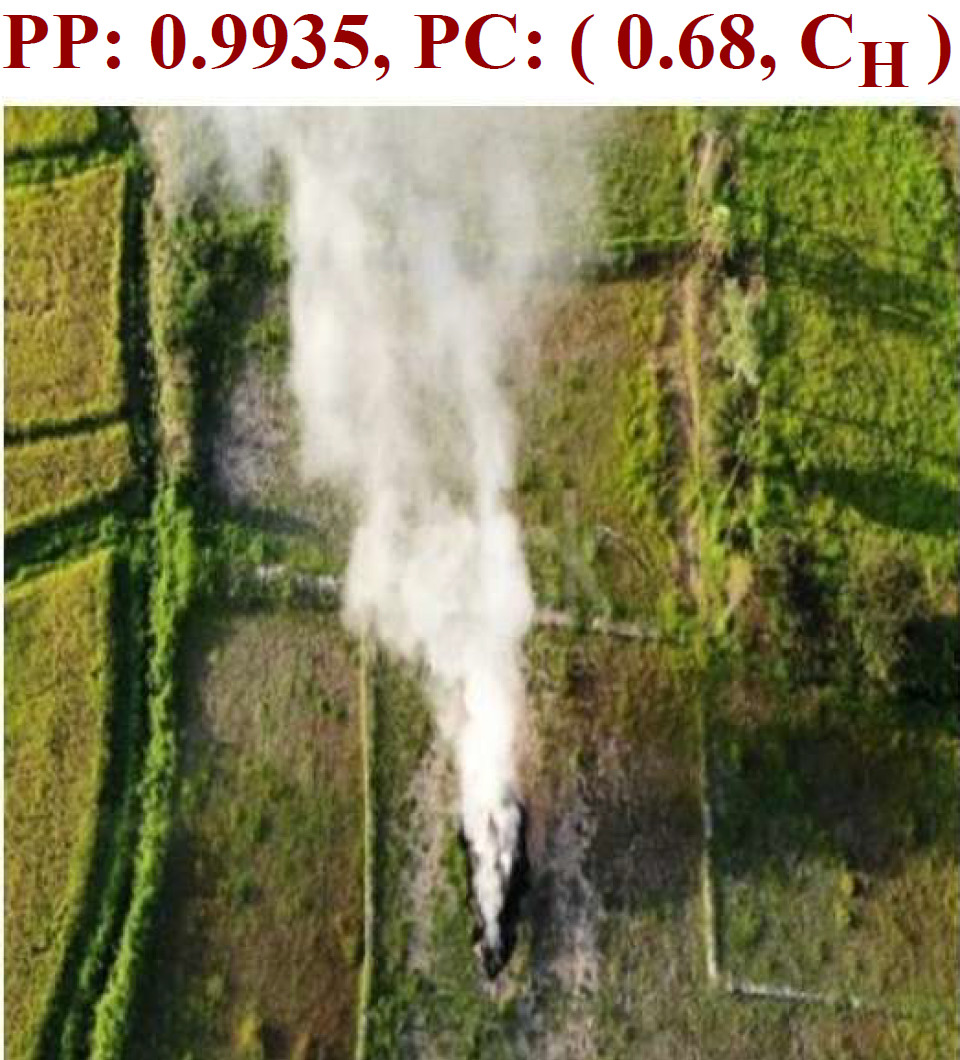} & \includegraphics[width=3.5cm, height=3.5cm]{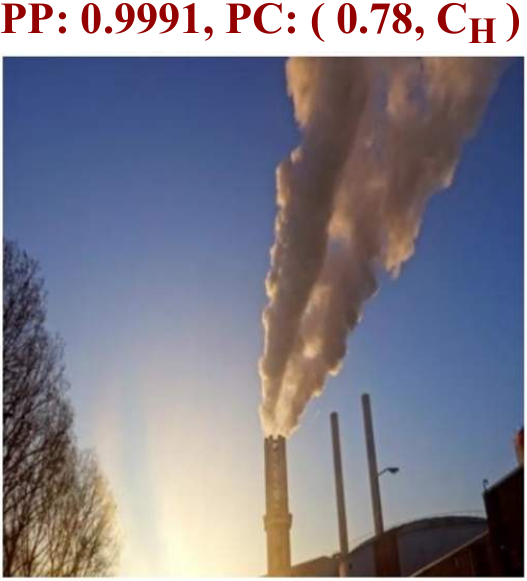}&\includegraphics[width=3.5cm, height=3.5cm]{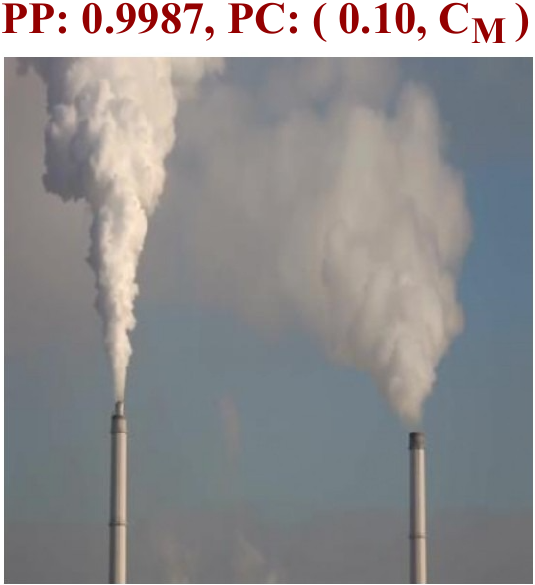}\\
 \textbf{Smoke1}&\textbf{Smoke2}&\textbf{Smoke3}&\textbf{Smoke5}\\
 \\
 \includegraphics[width=3.5cm, height=3.5cm]{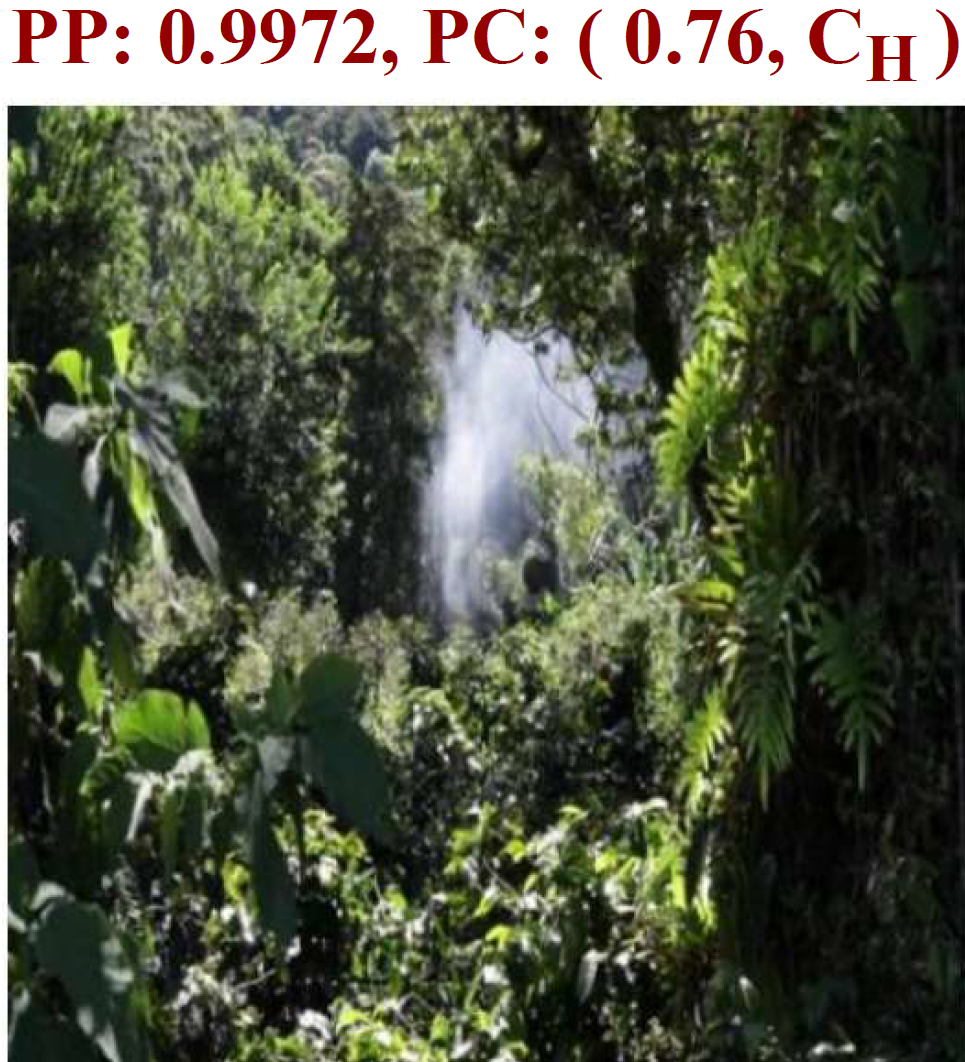}&\includegraphics[width=3.5cm, height=3.5cm]{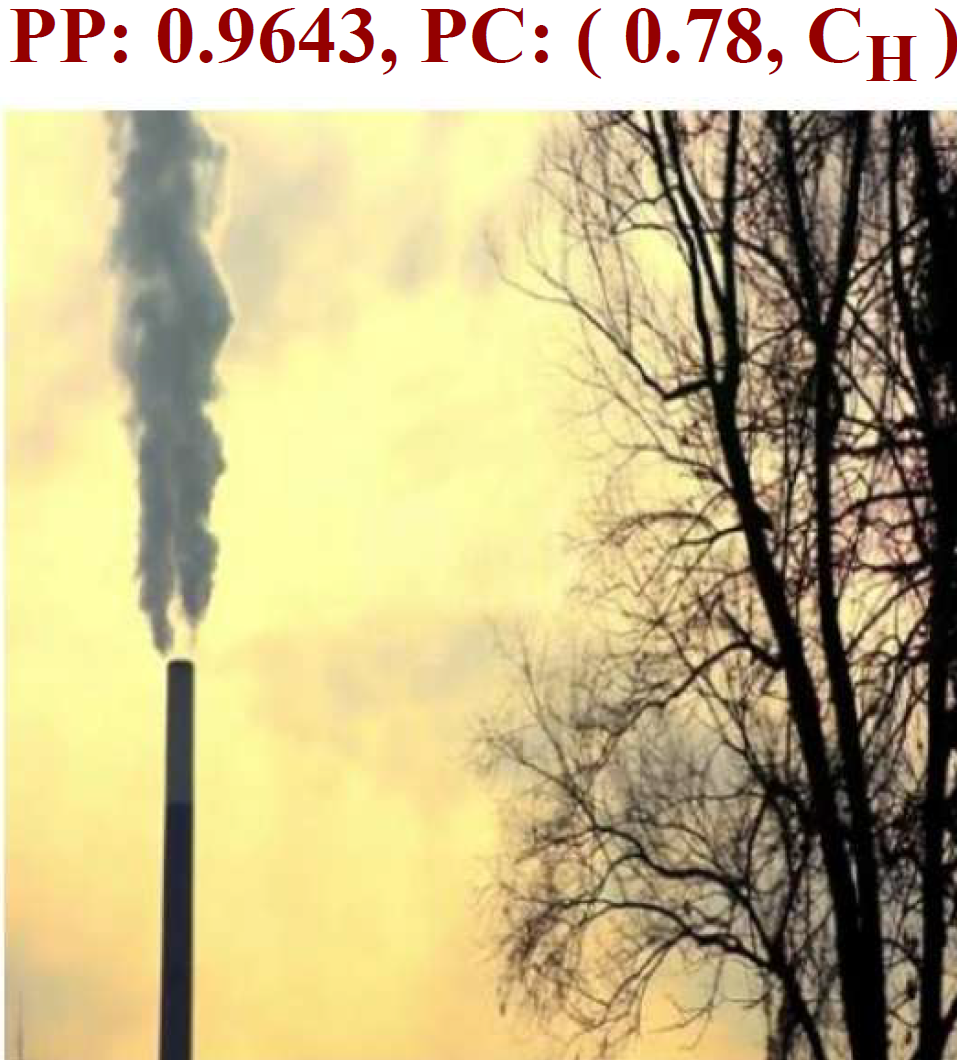}&\includegraphics[width=3.5cm, height=3.5cm]{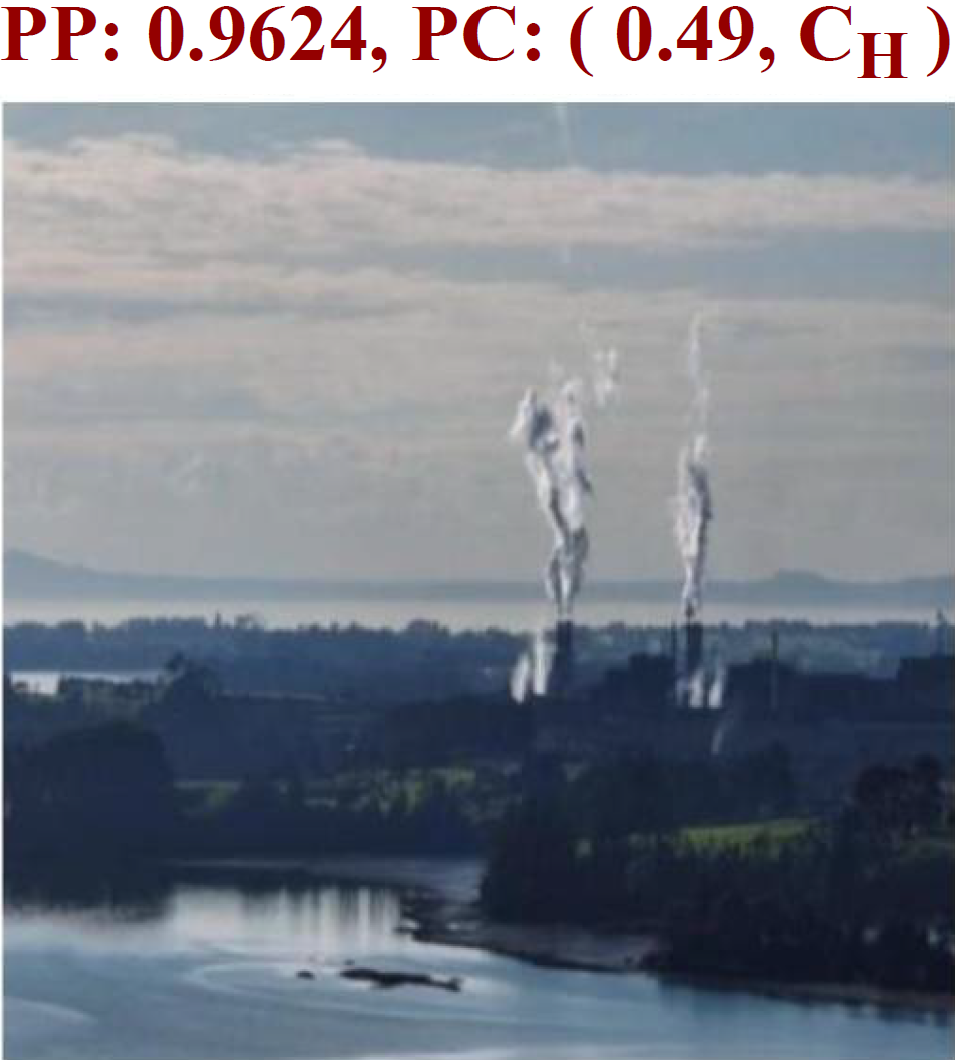}&\includegraphics[width=3.5cm, height=3.5cm]{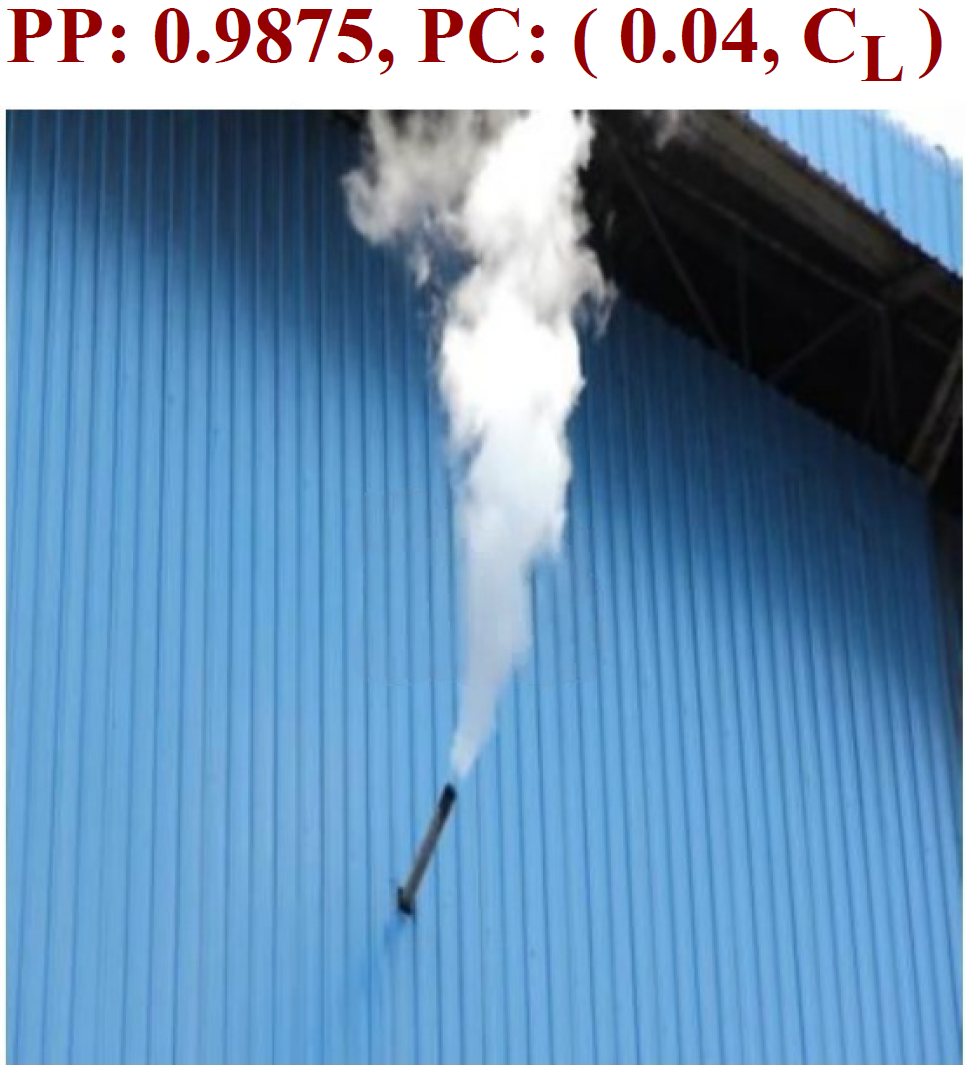}\\
 \textbf{Smoke6}&\textbf{Smoke7}&\textbf{Smoke8}&\textbf{Smoke9}\\
 \\
 \includegraphics[width=3.5cm, height=3.5cm]{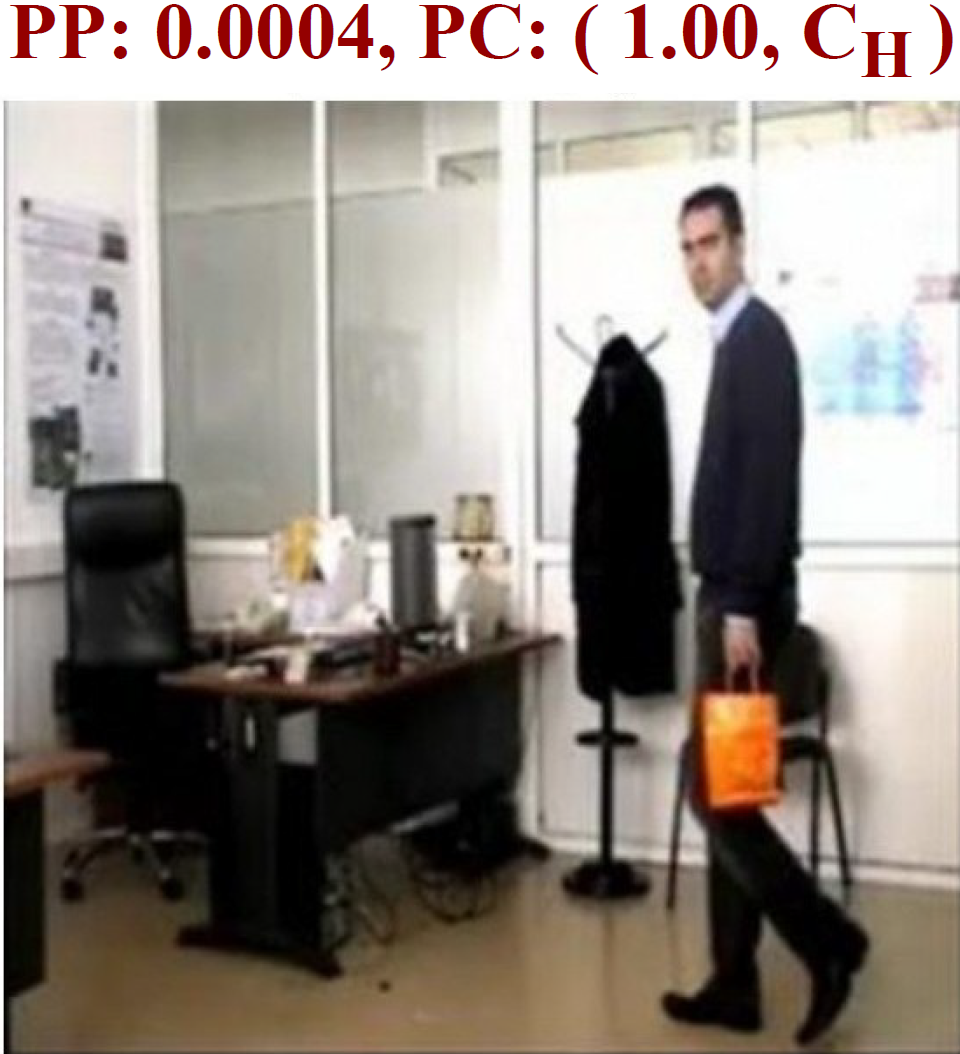} & \includegraphics[width=3.5cm, height=3.5cm]{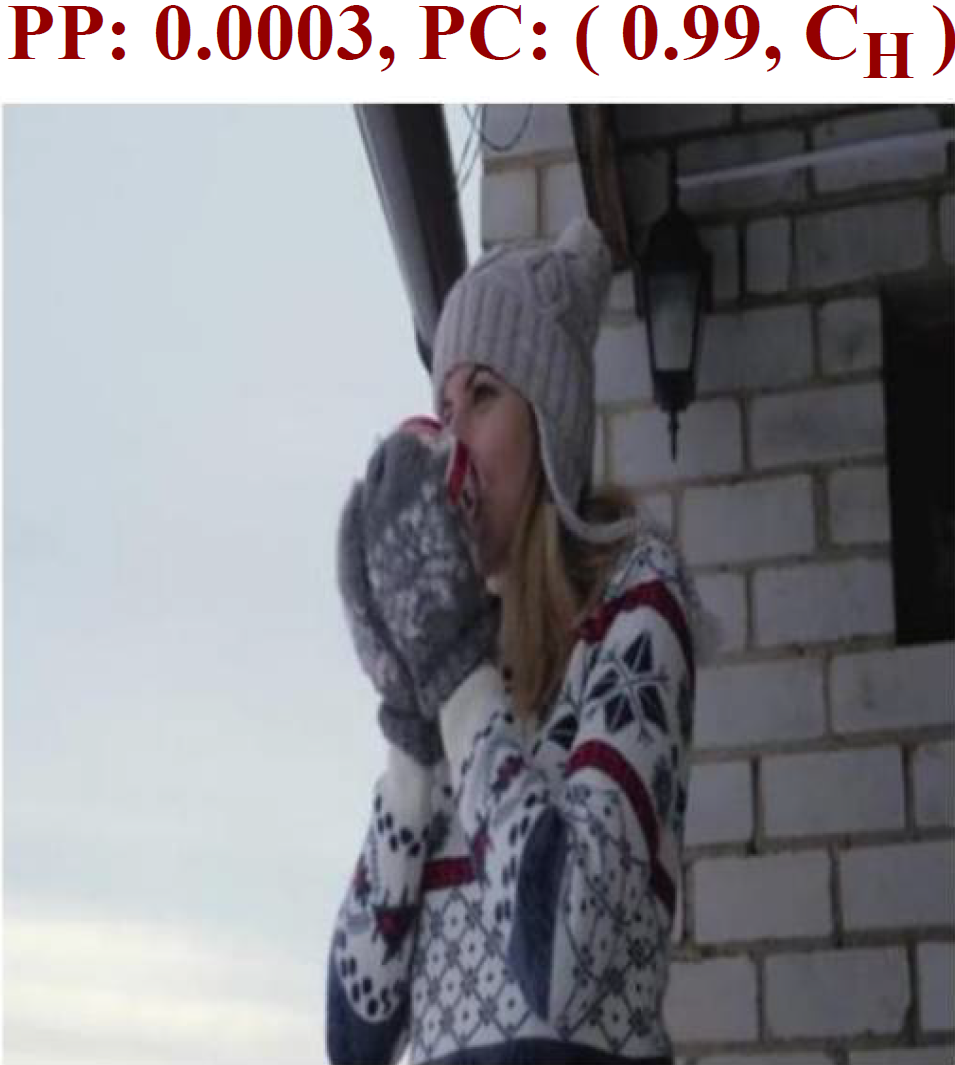} & \includegraphics[width=3.5cm, height=3.5cm]{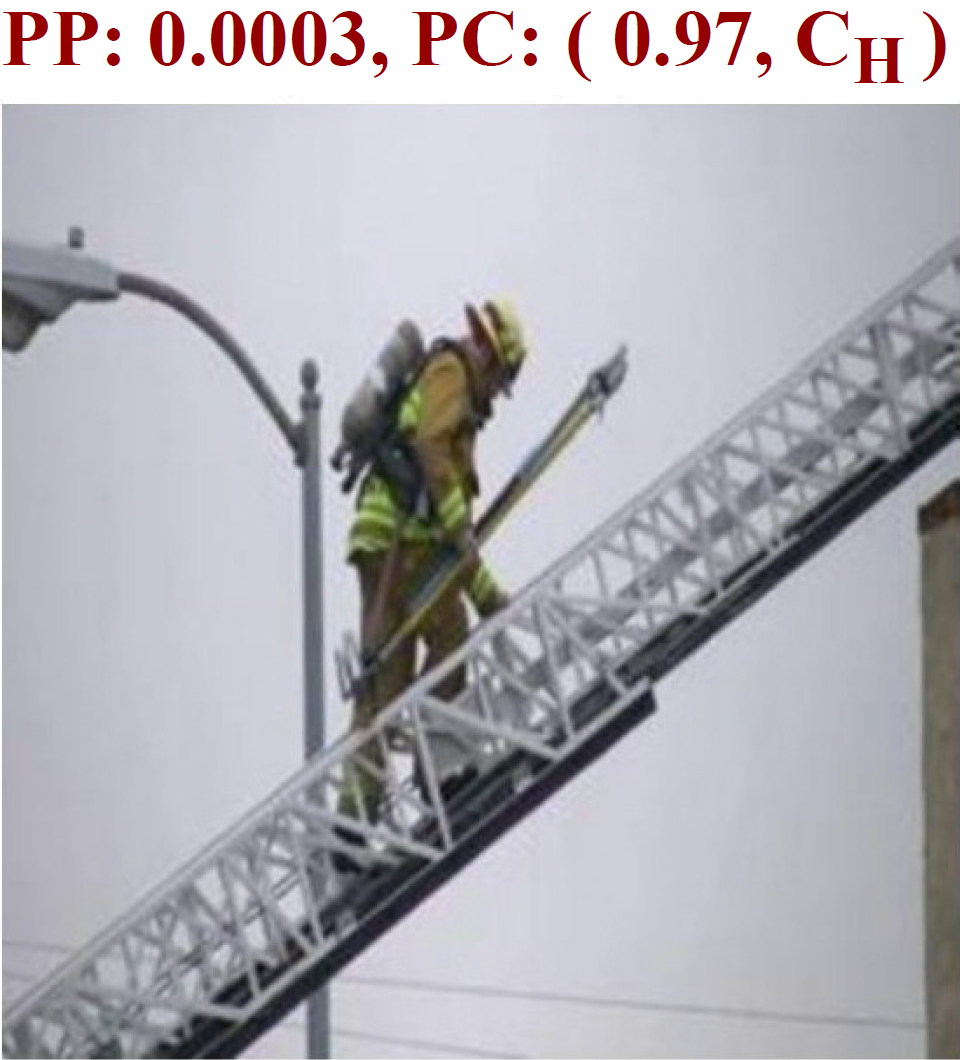} & \includegraphics[width=3.5cm, height=3.5cm]{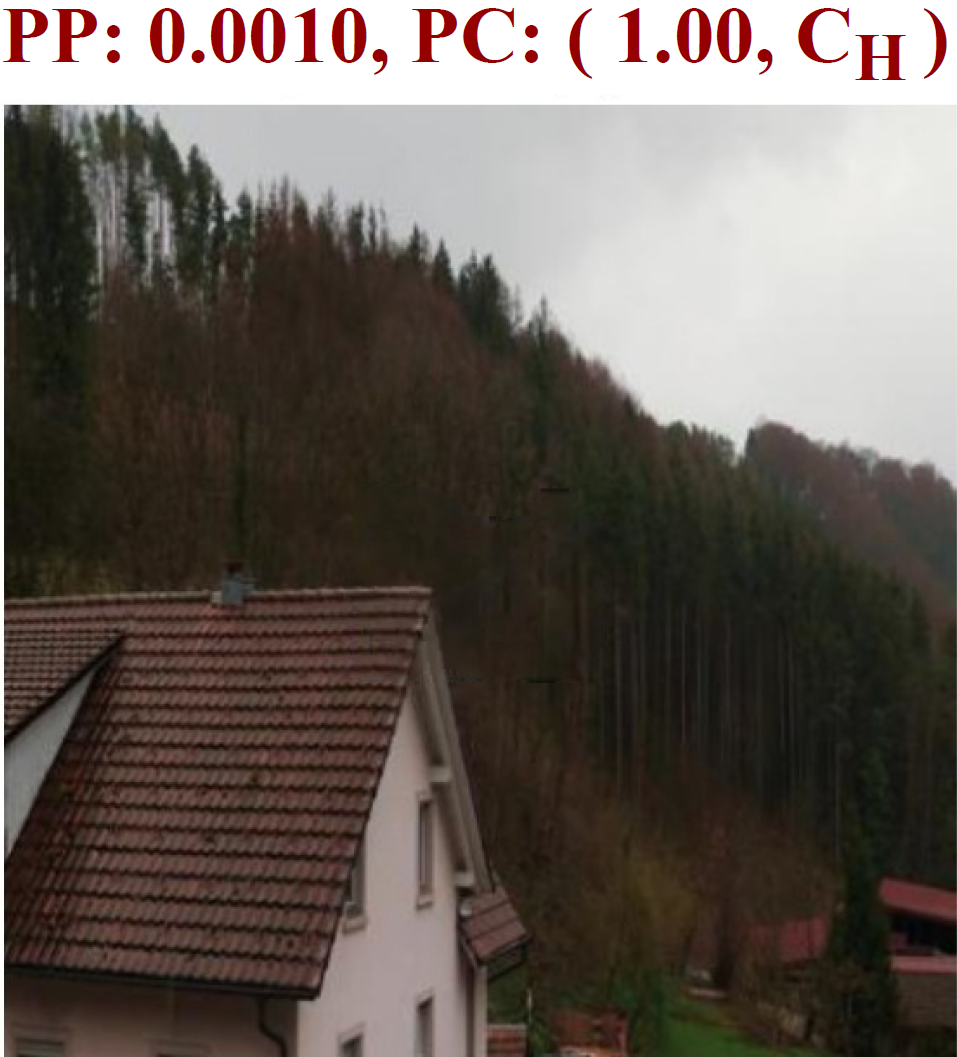}\\
 \textbf{Non-Smoke3} & \textbf{Non-Smoke5} & \textbf{Non-Smoke6} & \textbf{Non-Smoke7}\\
 \\
 \includegraphics[width=3.5cm, height=3.5cm]{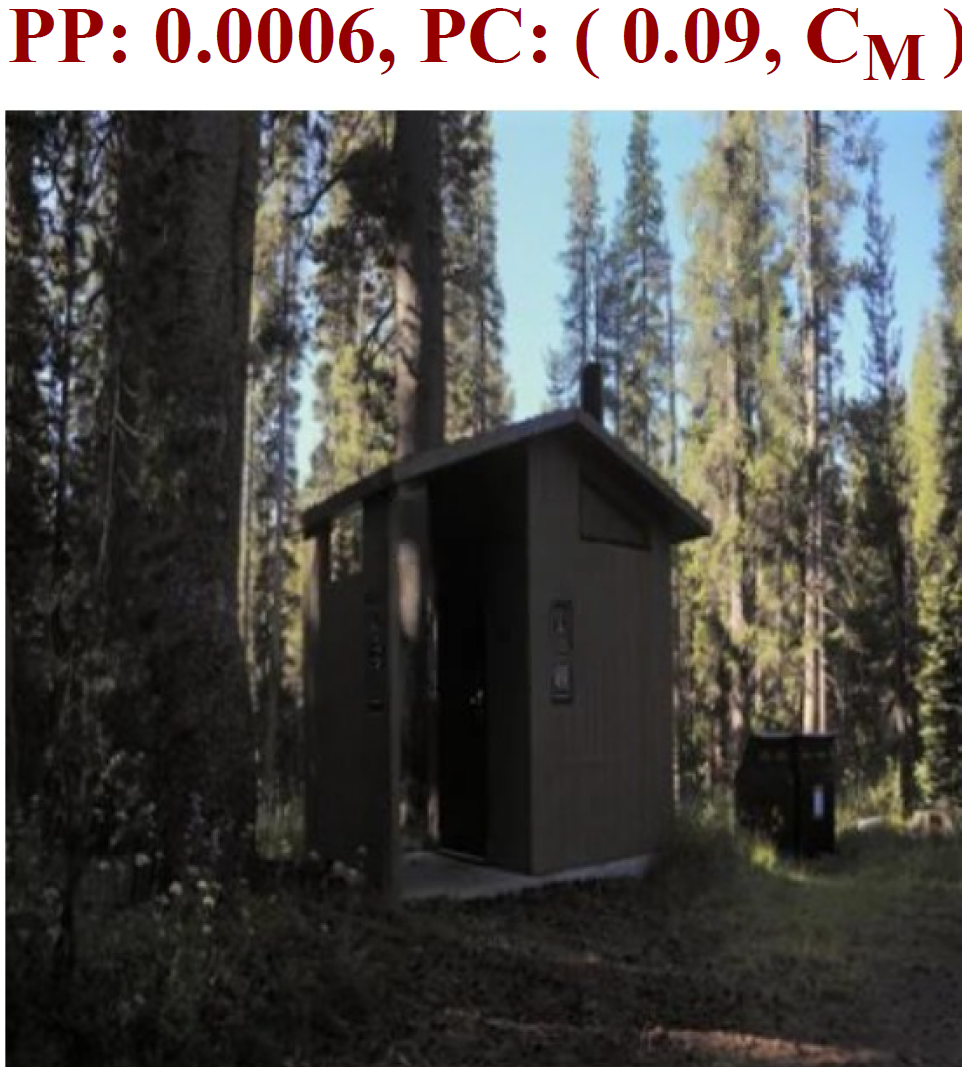} & \includegraphics[width=3.5cm, height=3.5cm]{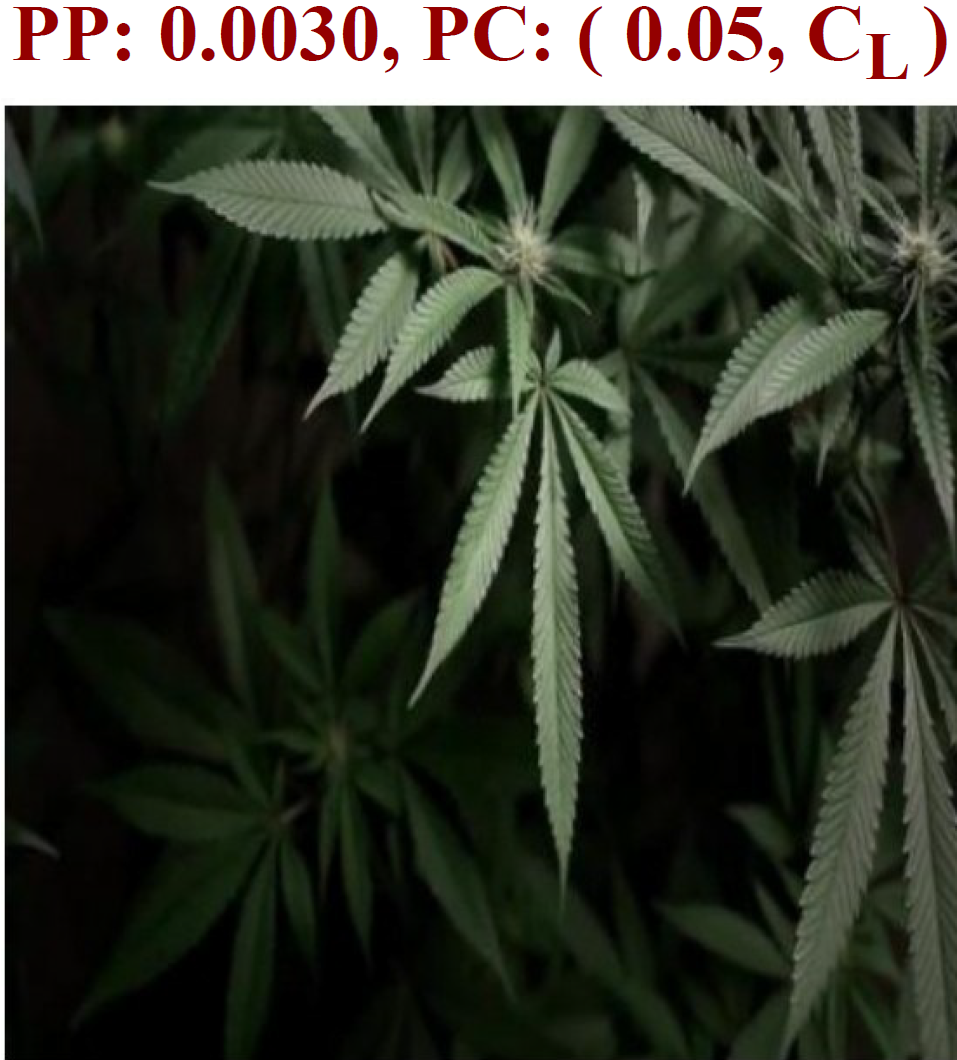} & \includegraphics[width=3.5cm, height=3.5cm]{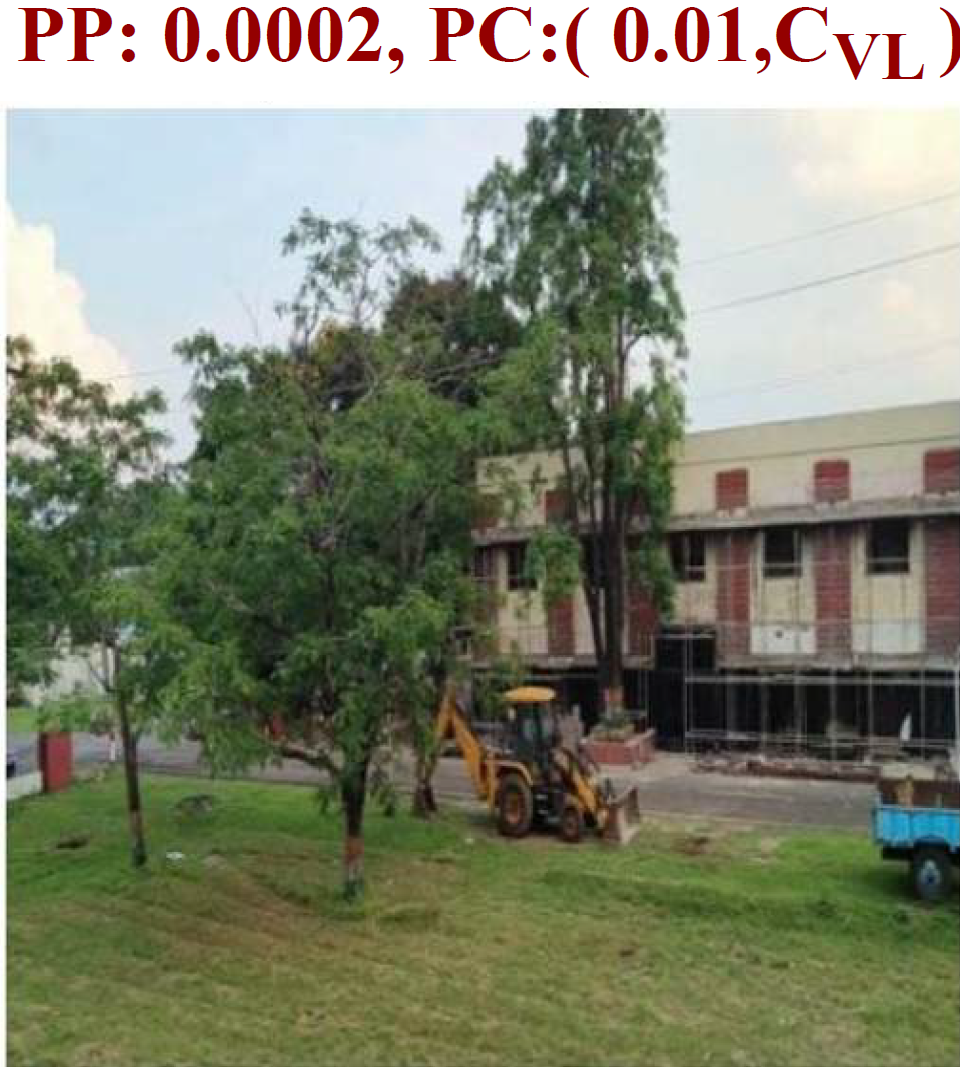} & \includegraphics[width=3.5cm, height=3.5cm]{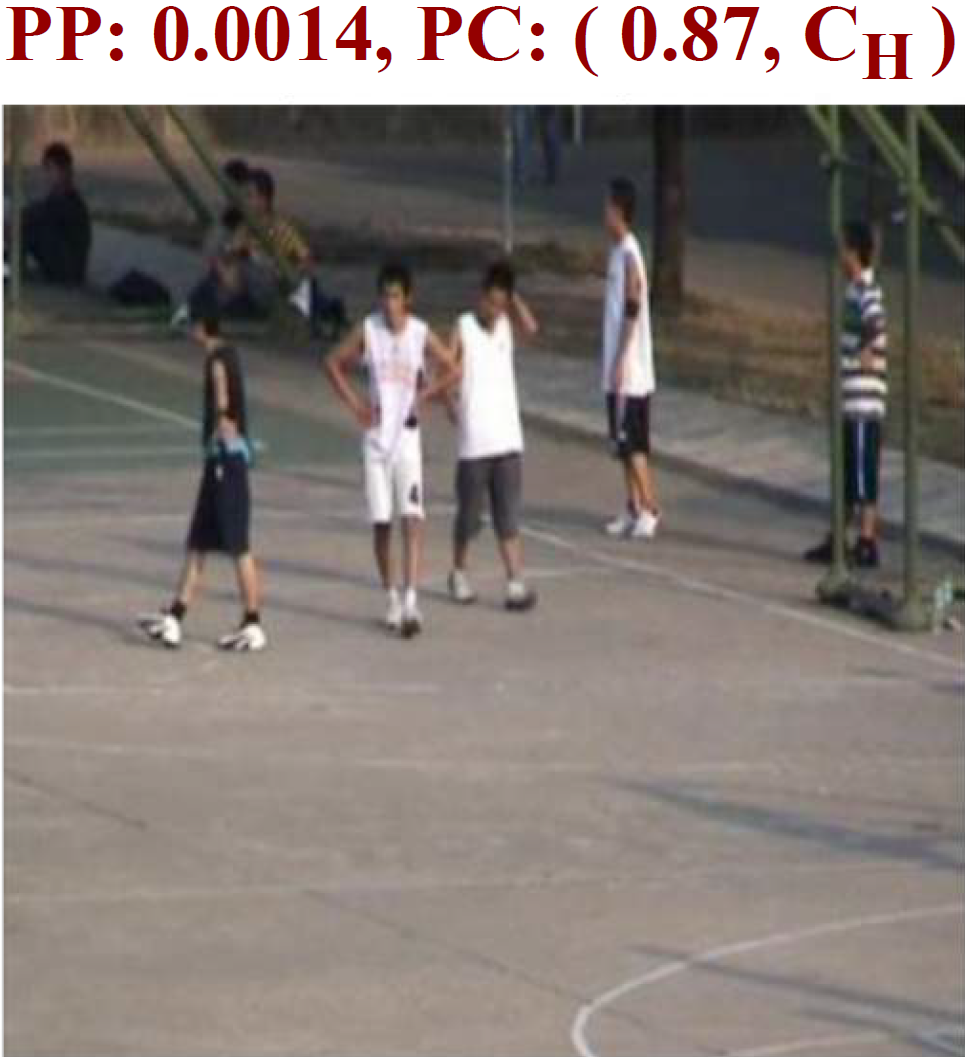}\\
 \textbf{Non-Smoke8} &\textbf{Non-Smoke9} & \textbf{Non-Smoke10} & \textbf{Non-Smoke11}\\
\end{tabular}
\caption{Results demonstrating the predicted probability ($PP$) and associated probability confidence ($PC$) scores corresponding to smoke samples (first and second row) and non-smoke samples (third and fourth row) from the TP-UAST model.}
\label{PSS}
\end{center}
\end{figure}

 The final experiment illustrates test-case outputs from the proposed TP-UAST model, as shown in Fig.~\ref{PSS}, reporting the predicted probability (PP) and its associated probability confidence (PC) score for each image in the smoke and non-smoke datasets. Across eight smoke examples, the network reliably identifies smoke. For instance, Smoke1, which contains a small campfire plume, yields PP=0.9993 with PC=0.68, indicating high confidence ($C_{H}$). Likewise, the dense emissions in Smoke2, Smoke3, and Smoke6-Smoke8 produce equally high confidence scores. By contrast, the more diffuse, cloud-like plumes in Smoke5 and Smoke9 return \(PC=0.10\) (moderate confidence ($C_{M}$)) and \(PC=0.04\) (low confidence ($C_{L}$)), respectively, underscoring the difficulty of discriminating thin, wispy smoke against complex terrain or sky-like backgrounds.
\par In the non-smoke dataset, all eight samples yield significantly low PP values, confirming strong negative discrimination. For instance, the indoor office scene in Non-Smoke3, the exhaled vapor sample of Non-Smoke5, the firefighter ascending a ladder in Non-Smoke6, the distant hillside view of Non-Smoke7, and the outdoor sports scene in Non-Smoke11 demonstrate high confidence associated PP values. While, challenging low-light scenarios, such as a wooded cabin in a dark forest setting of Non-Smoke8 and overlapping foliage edges in Non-Smoke9, produce \(PP=0.0006\), \(PC=0.09\) ($C_{M}$) and \(PP=0.0030\), \(PC=0.05\) ($C_{L}$). Cloud formations in the top-left and top-right regions of Non-Smoke10 yield \(PP=0.0002\), \(PC=0.01\), returning a very low confidence ($C_{VL}$) in the corresponding PP values. These results demonstrate that by jointly predicting \(PP\) and \(PC\), the model offers transparent insight into each decision, thereby ensuring reliable performance across diverse real-world environments.
\section{Conclusion}
In this work, a unified framework is introduced for smoke detection that fuses motion and appearance cues via an optical-flow-driven segmentation pipeline and a novel TP-UAST model. The proposed FCDLe-FOV model produces high-fidelity optical flow maps under variable illumination while preserving motion boundaries, and a Gaussian Mixture Model generates precise smoke region masks for training. The TP-UAST architecture leverages a dual-branch Swin Transformer backbone augmented with dedicated heads for point prediction, mean estimation, and variance regression. A two-phase curriculum decouples classification accuracy from uncertainty learning, yielding a classifier that achieves excellent detection metrics on the test set and produces well-calibrated confidence scores. Extensive quantitative and qualitative experiments demonstrate the robustness, reliability, and interpretability of the proposed approach across diverse scenarios. By explicitly modeling both aleatoric and epistemic uncertainties, TP-UAST not only identifies smoke with high precision and recall, but also quantifies its own confidence, enabling risk-aware decision thresholds in safety-critical applications. Future work will focus on real-time deployment of TP-UAST on embedded platforms with limited compute resources. Additionally, we plan to extend the framework to multi-class hazard detection (e.g., fire, steam, dust) to broaden its applicability in industrial and environmental monitoring.
\clearpage
\bibliographystyle{unsrt} 
\bibliography{main.bib}

\begin{thebibliography}{10}

\bibitem{Cui2020}
Fengmei Cui.
\newblock Deployment and integration of smart sensors with iot devices
  detecting fire disasters in huge forest environment.
\newblock {\em Computer Communications}, 150:818--827, 2020.

\bibitem{Liu2023}
Pingshan Liu, Pingchuan Xiang, and Dianjie Lu.
\newblock A new multi-sensor fire detection method based on lstm networks with
  environmental information fusion.
\newblock {\em Neural Computing and Applications}, 35(36):25275--25289, 2023.

\bibitem{Jin2025}
Peixian Jin, Pengle Cheng, Xiaodong Liu, and Ying Huang.
\newblock From smoke to fire: A forest fire early warning and risk assessment
  model fusing multimodal data.
\newblock {\em Engineering Applications of Artificial Intelligence},
  152:110848, 2025.

\bibitem{Bardeji2017}
Somayeh~Gh Bardeji, Isabel~N Figueiredo, and Erc{\'\i}lia Sousa.
\newblock Optical flow with fractional order regularization: variational model
  and solution method.
\newblock {\em Applied Numerical Mathematics}, 114:188--200, 2017.

\bibitem{Muzammil2023}
Khan Muzammil and Kumar Pushpendra.
\newblock A level set based fractional order variational model for motion
  estimation in application oriented spectrum.
\newblock {\em Expert Systems with Applications}, 219:119628, 2023.

\bibitem{Chambolle2004}
Antonin Chambolle.
\newblock An algorithm for total variation minimization and applications.
\newblock {\em Journal of Mathematical Imaging and Vision}, 20:89--97, 2004.

\bibitem{Farnoush2008}
R~Farnoush and PAK~B ZAR.
\newblock Image segmentation using gaussian mixture model.
\newblock 2008.

\bibitem{Mody2024}
Prerak Mody, Nicolas~F Chaves-de Plaza, Chinmay Rao, Eleftheria Astrenidou,
  Mischa de~Ridder, Nienke Hoekstra, Klaus Hildebrandt, and Marius Staring.
\newblock Improving uncertainty-error correspondence in deep bayesian medical
  image segmentation.
\newblock {\em arXiv preprint arXiv:2409.03470}, 2024.

\bibitem{Khan2023}
Muzammil Khan, Pushpendra Kumar, and Nitish~Kumar Mahala.
\newblock Cnn-based fire prediction using fractional order optical flow and
  smoke features.
\newblock In {\em Applications of Optimization and Machine Learning in Image
  Processing and IoT}, pages 156--180. Chapman and Hall/CRC, 2023.

\bibitem{Huo2022}
Yinuo Huo, Qixing Zhang, Yang Jia, Dongcai Liu, Jinfu Guan, Gaohua Lin, and
  Yongming Zhang.
\newblock A deep separable convolutional neural network for multiscale
  image-based smoke detection.
\newblock {\em Fire Technology}, pages 1--24, 2022.

\bibitem{Vicente2002}
Jerome Vicente and Philippe Guillemant.
\newblock An image processing technique for automatically detecting forest
  fire.
\newblock {\em International Journal of Thermal Sciences}, 41(12):1113--1120,
  2002.

\bibitem{Gu2019}
Ke~Gu, Zhifang Xia, Junfei Qiao, and Weisi Lin.
\newblock Deep dual-channel neural network for image-based smoke detection.
\newblock {\em IEEE Transactions on Multimedia}, 22(2):311--323, 2019.

\bibitem{Li2020}
Pu~Li and Wangda Zhao.
\newblock Image fire detection algorithms based on convolutional neural
  networks.
\newblock {\em Case Studies in Thermal Engineering}, 19:100625, 2020.

\bibitem{Lin2019}
Gaohua Lin, Yongming Zhang, Gao Xu, and Qixing Zhang.
\newblock Smoke detection on video sequences using 3d convolutional neural
  networks.
\newblock {\em Fire Technology}, 55:1827--1847, 2019.

\bibitem{Cheng2023}
Guangtao Cheng, Yancong Zhou, Shan Gao, Yingyu Li, and Hao Yu.
\newblock Convolution-enhanced vision transformer network for smoke
  recognition.
\newblock {\em Fire Technology}, 59(2):925--948, 2023.

\bibitem{Song2024}
Huajun Song and Yulin Chen.
\newblock Video smoke detection method based on cell root--branch structure.
\newblock {\em Signal, Image and Video Processing}, pages 1--9, 2024.

\bibitem{Safarov2024}
Furkat Safarov, Shakhnoza Muksimova, Misirov Kamoliddin, and Young~Im Cho.
\newblock Fire and smoke detection in complex environments.
\newblock {\em Fire}, 7(11):389, 2024.

\bibitem{Mardani2023}
Konstantina Mardani, Nicholas Vretos, and Petros Daras.
\newblock Transformer-based fire detection in videos.
\newblock {\em Sensors}, 23(6):3035, 2023.

\bibitem{Mueller2013}
Martin Mueller, Peter Karasev, Ivan Kolesov, and Allen Tannenbaum.
\newblock Optical flow estimation for flame detection in videos.
\newblock {\em IEEE Transactions on Image Processing}, 22(7):2786--2797, 2013.

\bibitem{Pundir2017}
Arun~Singh Pundir and Balasubramanian Raman.
\newblock Deep belief network for smoke detection.
\newblock {\em Fire technology}, 53:1943--1960, 2017.

\bibitem{Kumar2016a}
Pushpendra Kumar and Sanjeev Kumar.
\newblock A modified variational functional for estimating dense and
  discontinuity preserving optical flow in various spectrum.
\newblock {\em AEU-International Journal of Electronics and Communications},
  70(3):289--300, 2016.

\bibitem{Huang2020}
Zhenghua Huang and Aimin Pan.
\newblock Non-local weighted regularization for optical flow estimation.
\newblock {\em Optik}, 208:164069, 2020.

\bibitem{khondaker2020}
Arnisha Khondaker, Arman Khandaker, and Jia Uddin.
\newblock Computer vision-based early fire detection using enhanced chromatic
  segmentation and optical flow analysis technique.
\newblock {\em International Arab Journal of Information Technology},
  17(6):947--953, 2020.

\bibitem{Khan2024}
Muzammil Khan, Nitish~Kumar Mahala, and Pushpendra Kumar.
\newblock Caputo derivative based nonlinear fractional order variational model
  for motion estimation in various application oriented spectrum.
\newblock {\em S{\=a}dhan{\=a}}, 49(1):1--28, 2024.

\bibitem{Chunyu2010}
Yu~Chunyu, Fang Jun, Wang Jinjun, and Zhang Yongming.
\newblock Video fire smoke detection using motion and color features.
\newblock {\em Fire technology}, 46:651--663, 2010.

\bibitem{Wu2021}
Yuanlu Wu, Minghao Chen, Yan Wo, and Guoqiang Han.
\newblock Video smoke detection base on dense optical flow and convolutional
  neural network.
\newblock {\em Multimedia Tools and Applications}, 80:35887--35901, 2021.

\bibitem{Kikuta2024}
Kazutaka Kikuta, Ken~T Murata, and Yuki Murakami.
\newblock A daytime smoke detection method based on variances of optical flow
  and characteristics of hsv color on footage from outdoor camera in urban
  city.
\newblock {\em Fire Technology}, 60(3):1427--1452, 2024.

\bibitem{vese2002}
Luminita~A Vese and Tony~F Chan.
\newblock A multiphase level set framework for image segmentation using the
  mumford and shah model.
\newblock {\em International journal of computer vision}, 50:271--293, 2002.

\bibitem{Neelan2023}
Arun~Govind Neelan.
\newblock Von neumann stability analysis for multi-level multi-step methods.
\newblock {\em arXiv preprint arXiv:2310.08274}, 2023.

\bibitem{liu2021}
Ze~Liu, Yutong Lin, Yue Cao, Han Hu, Yixuan Wei, Zheng Zhang, Stephen Lin, and
  Baining Guo.
\newblock Swin transformer: Hierarchical vision transformer using shifted
  windows.
\newblock In {\em Proceedings of the IEEE/CVF international conference on
  computer vision}, pages 10012--10022, 2021.

\bibitem{Vaswani2017}
Ashish Vaswani, Noam Shazeer, Niki Parmar, Jakob Uszkoreit, Llion Jones,
  Aidan~N Gomez, Lukasz Kaiser, and Illia Polosukhin.
\newblock Attention is all you need. corr abs/1706.03762, 2017.

\bibitem{dong2023}
Haitao Dong, Chengjun Chen, Jinlei Wang, Feixiang Shen, and Yong Pang.
\newblock Vit-saps: Detail-aware transformer for mechanical assembly semantic
  segmentation.
\newblock {\em IEEE Access}, 11:41467--41479, 2023.

\bibitem{Raisi2020}
Zobeir Raisi, Mohamed~A Naiel, Paul Fieguth, Steven Wardell, and John Zelek.
\newblock 2d positional embedding-based transformer for scene text recognition.
\newblock {\em Journal of Computational Vision and Imaging Systems}, 6(1):1--4,
  2020.

\bibitem{Lu2019a}
Jin Lu, Hua Yang, Qinghu Zhang, and Zhouping Yin.
\newblock A field-segmentation-based variational optical flow method for piv
  measurements of nonuniform flows.
\newblock {\em Experiments in Fluids}, 60:1--17, 2019.

\bibitem{Liu2025a}
Hongying Liu, Fuquan Zhang, Yiqing Xu, Junling Wang, Hong Lu, Wei Wei, and Jun
  Zhu.
\newblock Tfnet: Transformer-based multi-scale feature fusion forest fire image
  detection network.
\newblock {\em Fire}, 8(2):59, 2025.

\bibitem{Yang2023}
Huanyu Yang, Jun Wang, and Jiacun Wang.
\newblock Efficient detection of forest fire smoke in uav aerial imagery based
  on an improved yolov5 model and transfer learning.
\newblock {\em Remote Sensing}, 15(23):5527, 2023.

\bibitem{Almeida2022}
Jefferson~Silva Almeida, Chenxi Huang, Fabr{\'\i}cio~Gonzalez Nogueira, Surbhi
  Bhatia, and Victor Hugo~C de~Albuquerque.
\newblock Edgefiresmoke: A novel lightweight cnn model for real-time video
  fire--smoke detection.
\newblock {\em IEEE Transactions on Industrial Informatics}, 18(11):7889--7898,
  2022.

\bibitem{Saponara2021}
Sergio Saponara, Abdussalam Elhanashi, and Alessio Gagliardi.
\newblock Real-time video fire/smoke detection based on cnn in antifire
  surveillance systems.
\newblock {\em Journal of Real-Time Image Processing}, 18:889--900, 2021.

\bibitem{He2016}
Kaiming He, Xiangyu Zhang, Shaoqing Ren, and Jian Sun.
\newblock Deep residual learning for image recognition.
\newblock In {\em IEEE conference on computer vision and pattern recognition},
  pages 770--778, 2016.

\bibitem{Simonyan2014}
Karen Simonyan and Andrew Zisserman.
\newblock Very deep convolutional networks for large-scale image recognition.
\newblock {\em arXiv preprint arXiv:1409.1556}, 2014.

\bibitem{Wang2025}
Yupeng Wang, Yongli Wang, Zaki~Ahmad Khan, Anqi Huang, and Jianghui Sang.
\newblock Multi-level feature fusion networks for smoke recognition in remote
  sensing imagery.
\newblock {\em Neural Networks}, 184:107112, 2025.

\bibitem{Kong2024}
Derui Kong, Yinfeng Li, and Manzhen Duan.
\newblock Fire and smoke real-time detection algorithm for coal mines based on
  improved yolov8s.
\newblock {\em Plos one}, 19(4):e0300502, 2024.

\bibitem{Sozol2025}
Md~Shafak~Shahriar Sozol, M~Rubaiyat~Hossain Mondal, and Achmad~Husni Thamrin.
\newblock Indoor fire and smoke detection based on optimized yolov5.
\newblock {\em PLoS One}, 20(4):e0322052, 2025.

\bibitem{Howard2017}
Andrew~G Howard, Menglong Zhu, Bo~Chen, Dmitry Kalenichenko, Weijun Wang,
  Tobias Weyand, Marco Andreetto, and Hartwig Adam.
\newblock Mobilenets: Efficient convolutional neural networks for mobile vision
  applications.
\newblock {\em arXiv preprint arXiv:1704.04861}, 2017.

\bibitem{Ballester2016}
Pedro Ballester and Ricardo Araujo.
\newblock On the performance of googlenet and alexnet applied to sketches.
\newblock In {\em Proceedings of the AAAI conference on artificial
  intelligence}, volume~30, 2016.

\end{thebibliography}
\end{document}